\documentclass{article} 
\usepackage{iclr2027_conference,times}

\usepackage{amsmath,amsfonts,bm}

\def\eqref#1{equation~\ref{#1}}

\def\1{\bm{1}}

\DeclareMathAlphabet{\mathsfit}{\encodingdefault}{\sfdefault}{m}{sl}
\SetMathAlphabet{\mathsfit}{bold}{\encodingdefault}{\sfdefault}{bx}{n}

\usepackage{hyperref}
\usepackage{url}
\usepackage{booktabs}
\usepackage{amssymb}  
\usepackage{wrapfig}
\usepackage{graphicx}
\usepackage{multirow}
\usepackage{xcolor}
\usepackage{tikz}
\usepackage{titletoc}
\usetikzlibrary{positioning,calc,arrows.meta}
\usepackage{subcaption}
\usepackage[T1]{fontenc}
\usepackage[table]{xcolor}
\definecolor{sharedrow}{HTML}{EEF3F9}
\definecolor{phaserow}{HTML}{FBF3E8}

\definecolor{lotusrow}{HTML}{EDE7F8}
\newcommand{\lc}{\cellcolor{lotusrow}}
\newcommand{\seqarrow}{\raisebox{0.3ex}{\rotatebox[origin=c]{180}{$\Lsh$}}}

\usepackage[most]{tcolorbox}
\usepackage{float}
\newtcolorbox{promptbox}[1]{enhanced, breakable,
  colback=gray!4, colframe=gray!50, boxrule=0.5pt, arc=2pt,
  left=5pt, right=5pt, top=4pt, bottom=4pt,
  title=#1, fonttitle=\bfseries\footnotesize, coltitle=black, colbacktitle=gray!14}
\newcommand{\ppart}[1]{{\rmfamily\color{gray!80!black}\scriptsize\textsc{#1}}\\[1pt]}
\newcommand{\tsblock}{\textcolor{violet!75!black}{<ts>\,\ldots\,</ts>}}

\newcommand{\draftnotes}{}
\ifdefined\draftnotes
  \newcommand{\todo}[1]{\textcolor{red}{\textbf{[TODO: #1]}}}
\else
  \newcommand{\todo}[1]{}
\fi

\title{Lotus: Learning to Understand Time-series by Learning to Forecast}

\author{Van Dai Do$^*$, Huu Hiep Nguyen$^*$, Minh Hoang Nguyen, Manh Nguyen and Hung Le \\
Deakin's Applied Artificial Intelligence Initiative, Australia \\
\texttt{\{v.do, huu.n, s223669184, manh.nguyen, thai.le\}@deakin.edu.au}
}

\providecommand{\method}{Ours}
\renewcommand{\method}{\textsc{Lotus}} 

\iclrfinalcopy 
\begin{document}

\maketitle
\lhead{Preprint}          
{\renewcommand{\thefootnote}{*}\footnotetext{Equal contribution.}}
\suppressfloats[t]

\begin{abstract}
Time-series language models typically align numerical signals with language using labels, captions, or question--answer pairs, without explicitly requiring their representations to preserve how the underlying signals evolve. We introduce \emph{forecast grounding}, a pretraining principle that requires language-aligned time-series representations to remain predictive of the future evolution before they are adapted to downstream tasks. We instantiate this principle in \method{}, which divides each time-series channel into non-overlapping patches and maps them into the LLM embedding space. It then predicts the next patch from the LLM hidden states of the time-series tokens, encouraging the learned representation to retain temporal information. To capture both temporal changes and periodic patterns, we also propose a \emph{temporal--spectral patch representation} that combines each raw time-series patch with its spectral representation before mapping it into the LLM embedding space. Across three language model backbones and six downstream evaluation datasets, \method{} achieves the best average performance with minimal trainable parameters, reaching \(59.91\%\) accuracy and outperforming the strongest matched baseline by \(3.36\%\). Our representation analyses further show that \method{} uses the numerical input and preserves temporal structure in its learned patch representations. These results establish forecast grounding as a promising pretraining principle, enabling compact time-series language models to outperform models 25–50× their size on TSRBench. Our code is publicly available at \textit{https://github.com/DA2I2-SLM/Lotus}.
\end{abstract}
\section{Introduction}
\label{sec:intro}

\begin{wrapfigure}{r}{0.45\textwidth}
\vspace{-\intextsep}
\centering
\includegraphics[width=0.45\textwidth]{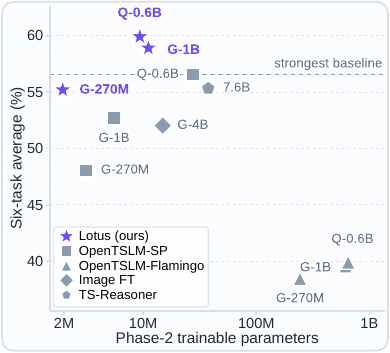}
\caption{\textbf{Six-task average exact match accuracy versus trainable parameters (LoRA).} G and Q denote Gemma3 and Qwen3 backbones, respectively.}
\label{fig:params-vs-performance}
\vspace{-\intextsep}
\end{wrapfigure}

Time series describe processes that evolve over time, making both the observed values and their temporal dynamics important for understanding the underlying signal. Time-series foundation models exploit this structure through objectives such as forecasting and reconstruction to learn transferable numerical representations~\citep{ansari2024chronos,das2024timesfm,woo2024moirai,nie2023patchtst}. At the same time, many practical tasks require understanding time series through natural language, such as identifying when an abnormal event occurred, comparing changes across variables, or answering questions about a physiological signal. Recent time-series language models address this setting by mapping numerical observations into representations that pretrained large language models (LLMs) can process~\citep{li2025sensorllm,cheng2025instructime,chatts,opentslm}. The effectiveness of this approach therefore depends on what information these time-series representations preserve.

Existing time-series language models are primarily aligned through language supervision, including class labels, captions, instructions, and question--answer pairs~\citep{opentslm, chatts}. These objectives encourage the representation to preserve information needed to produce the correct language output, but they do not explicitly require it to retain how the underlying signal may evolve. For example, two motion segments may both correspond to \emph{walking} while differing in speed, local dynamics, and likely continuation. The same language target may therefore correspond to temporal states with different future behavior. We argue that forecasting provides a natural additional constraint for learning such representations. Based on this idea, we introduce \textbf{forecast grounding}: during pretraining, a language-aligned time-series representation is also required to predict the future evolution of its source signal. If two observations share the same language target but lead to different futures, the forecasting objective provides an additional learning signal that distinguishes them. Forecasting therefore serves as a pretraining constraint that encourages the representation supplied to the LLM to preserve future-relevant temporal information, rather than as an additional downstream task.

We instantiate this principle in \method{}. Each channel is divided into non-overlapping patches and mapped by an encoder and projector to soft tokens in the LLM embedding space. Temporal and channel positions identify the structure of multivariate inputs, while the causal LLM provides interaction across signal patches and text. During pretraining, language examples train the model to generate answers, while forecasting examples train it to predict the next patch from the LLM states of the signal tokens. As a result, the forecasting loss directly shapes the representation the language model uses. After pretraining, the model is jointly adapted to different downstream time-series tasks. Beyond the pretraining objective, we also consider how local time-series information is represented before entering the LLM. We propose a \emph{temporal--spectral patch representation} that combines the raw temporal patch with its Fourier transform before mapping it into the LLM embedding space. The temporal view preserves local values and changes, while the spectral view exposes frequency content within each patch. Longer-range dependencies across patches are subsequently modeled by the language model. We evaluate \method{} with three language backbones on six downstream splits covering activity recognition, sleep staging, ECG question answering, and general time-series question answering. As shown in Figure~\ref{fig:params-vs-performance}, \method{} achieves the best average accuracy in every matched backbone group while using only a small number of trainable parameters.

Our contributions are threefold:
\begin{enumerate}
\item We propose \textbf{forecast grounding}, which uses future prediction during pretraining to encourage language-aligned time-series representations to preserve future-relevant temporal information.
\item We propose a \textbf{temporal--spectral patch representation} that combines the raw time-series patch with its spectral representation before mapping it into the LLM embedding space. This allows the model to capture both local temporal changes and frequency patterns.
\item We evaluate \method{} with three language model backbones on six downstream splits. \method{} achieves the best average accuracy in every matched backbone group, demonstrating the effectiveness of combining forecast grounding with temporal--spectral patch representations for time-series language modeling.
\end{enumerate}

\section{Related work}
\label{sec:related}

\subsection{Time-series language models and reasoning}

Recent work extends pretrained language models to numerical time series and studies their ability to perform classification, interpretation, and reasoning over signals. SensorLLM aligns motion signals with textual trend descriptions before activity-recognition tuning, while InstructTime formulates time-series classification through multimodal language modeling~\citep{li2025sensorllm,cheng2025instructime}. ChatTS pairs synthetic multivariate signals with attribute descriptions and instructions for reasoning over temporal attributes and relationships, while Time-MQA develops large-scale question--answer data for numerical analysis and open-ended interaction~\citep{chatts,kong2025timemqa}. In physiological domains, OpenTSLM studies activity recognition, sleep staging, and ECG question answering, while GEM grounds language generation in measurable ECG characteristics~\citep{opentslm,lan2025gem}. TS-Reasoner instead incorporates representations from a pretrained time-series foundation model and aligns them with an LLM through instruction tuning~\citep{yu2026tsreasoner}.

Despite this progress, reasoning directly over numerical signals remains difficult. \citet{merrill2024language} show that strong LLMs remain unreliable when time series are represented directly as text. Existing approaches address this problem mainly by introducing stronger language supervision, task-specific instruction data, or more informative numerical features. These directions improve what the language model is taught or what numerical information it receives, but place less emphasis on how the underlying time-series representation is shaped before downstream reasoning. \method{} focuses on this representation. We encode each patch using both its temporal values and spectral information, providing the language model with complementary views of local signal behavior. We further use forecasting during pretraining to encourage the resulting representation to retain information about future signal evolution. The goal is therefore not to introduce another reasoning objective, but to learn a stronger numerical representation on which later language understanding and reasoning can operate.

\subsection{Time-series--language alignment}

A central challenge in time-series language modeling is mapping continuous numerical signals into representations that pretrained language models can effectively process. Existing methods adopt several alignment mechanisms. Projection-based approaches map encoded time-series segments directly into the LLM embedding space~\citep{li2025sensorllm,opentslm}, while cross-attention architectures retain separate signal representations that language states can attend to~\citep{opentslm,lan2025gem}. TS-Reasoner instead projects representations from a pretrained time-series foundation model into the language-model space~\citep{yu2026tsreasoner}. These approaches establish effective connections between numerical signals and language, but the resulting representations are primarily shaped through language or downstream task supervision.

A related line of work uses the time series itself as a learning signal. PatchTST learns representations through masked-patch reconstruction~\citep{nie2023patchtst}, while SimMTM reconstructs the original series from multiple masked views to preserve temporal variation~\citep{dong2023simmtm}. Prediction has also been explored in language-model-based time-series methods. Time-LLM reprograms time-series patches using text prototypes before passing them to a frozen LLM for forecasting~\citep{jin2024timellm}, while TEST aligns time-series embeddings with a text-prototype space for prediction with a frozen language model~\citep{sun2024test}. More closely related to our predictive objective, aLLM4TS uses causal next-patch and multi-patch prediction to learn transferable time-series representations with an LLM~\citep{bian2024allm4ts}. These methods demonstrate that reconstruction and prediction can provide useful supervision for learning temporal representations, but they are developed primarily for forecasting or conventional time-series tasks. \method{} brings predictive supervision into time-series--language alignment for downstream language tasks through \textbf{forecast grounding}. During pretraining, we predict future patches from the post-LLM signal states, so the forecasting objective directly constrains the representations that are later used for language understanding and reasoning. Together with the temporal--spectral patch representation, this encourages the language model to operate on signal representations that capture both local temporal dynamics and frequency-domain structure while retaining information relevant to future signal evolution.

\section{Methodology}
\label{sec:method}

\begin{figure*}[t]
    \centering
    \includegraphics[width=\textwidth]{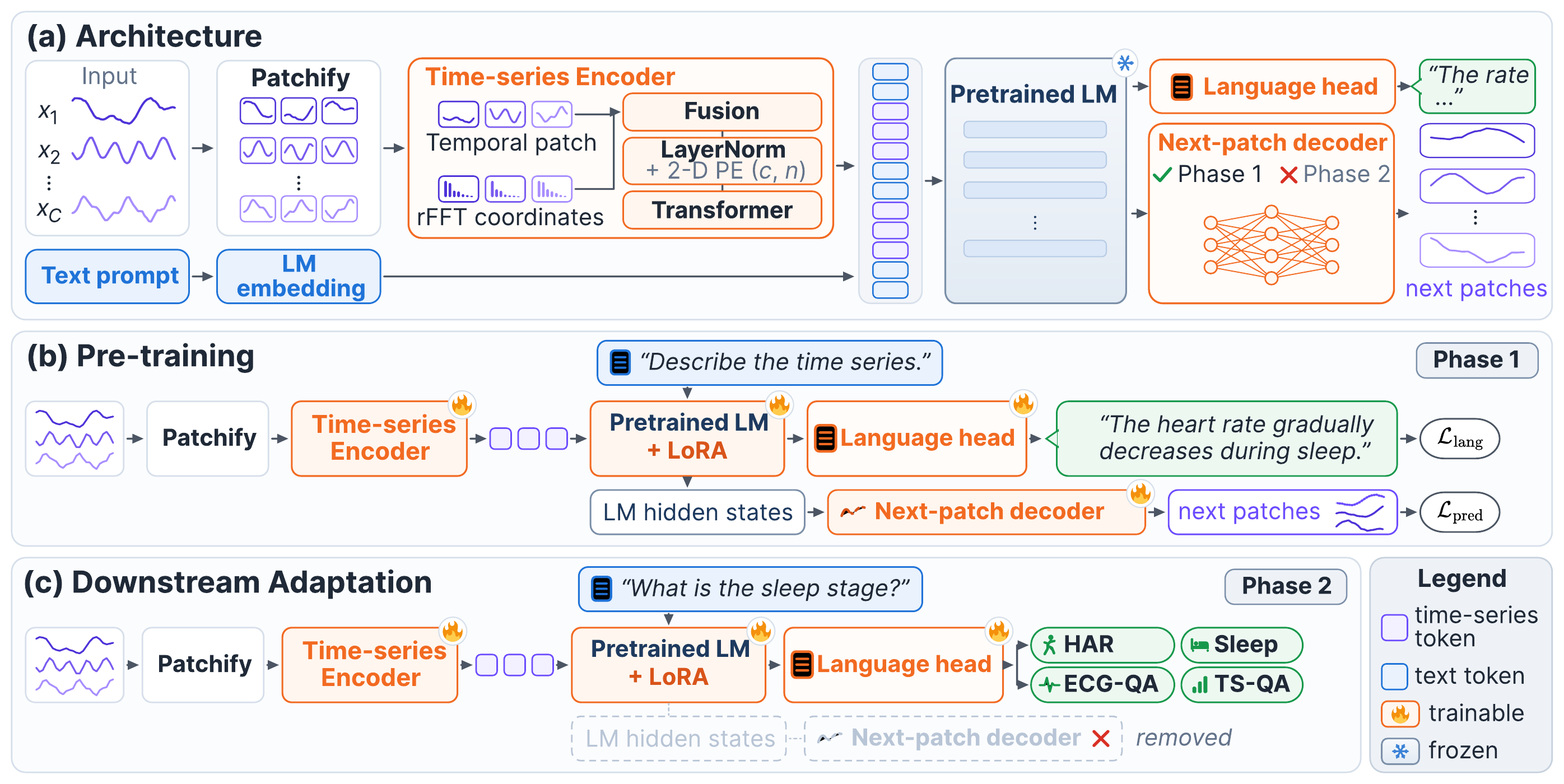}
    \caption{\textbf{Overview of \method{}.}
\textbf{(a) Architecture:} The architecture maps time-series patches to soft tokens that are processed jointly with text by the pretrained language model.
\textbf{(b) Pretraining:} During forecast grounding pretraining, language supervision trains textual outputs while next-patch prediction constrains LLM's hidden states to retain future-relevant information.
\textbf{(c) Downstream Adaptation:} During downstream adaptation, the next-patch decoder is removed and the model is trained with language supervision only.} \label{fig:overview}
\end{figure*}

Figure~\ref{fig:overview} summarizes \method{} in three parts. First, the architecture divides each time-series channel into patches and maps each patch to a soft time-series token in the language-model embedding space. This design allows the representation of each patch to be contextualized jointly with other patches and language. We use \emph{patch} to refer to a segment of the original time series and \emph{soft time-series token} to refer to its projected representation supplied to the language model. Second, forecast grounding applies predictive supervision to the language-model states corresponding to these soft time-series tokens. We hypothesize that requiring the same representations used for language processing to retain information about future time series provides a complementary training signal for downstream time-series understanding. Third, downstream adaptation removes the next-patch decoder and continues to adapt the time-series encoder and LLM using the language objective alone.

\subsection{\method{} Architecture}
\label{sec:arch}

\paragraph{Patchification.}
Let a multivariate time series be $\mathbf{S}\in\mathbb{R}^{C\times L}$, where $C$ is the number of channels and $L$ is the sequence length. Each channel is normalized to zero mean and unit variance over its observed, non-padded extent, zero-padded to a multiple of the patch length $P$, and divided into non-overlapping patches
\begin{equation}
    \mathbf{x}_{c,n}\in\mathbb{R}^{P},
    \qquad
    c\in\{1,\ldots,C\},
    \quad
    n\in\{1,\ldots,N\},
\end{equation}
where $N=\lceil L/P\rceil$. The input therefore yields $CN$ patch positions. The encoder operates on each patch independently, while interaction across patches and channels is handled later by the pretrained language model. We represent each patch in both temporal and Fourier coordinates to expose complementary views of the same local segment. The temporal view preserves the raw within-patch evolution, while the Fourier view provides an explicit frequency-aware parameterization that may facilitate learning periodic structure.

\paragraph{Time-Series Encoder.}
While the raw temporal patch directly preserves local values and short-term changes, patterns may be less explicit in the time domain. We therefore propose a \emph{temporal-spectral patch representation} which represents each patch using temporal and spectral views of its local dynamics. 
Specifically, for a patch $\mathbf{x}_{c,n}$, we use the raw patch as the temporal view. For the spectral view, let $\mathcal{F}(\mathbf{x}_{c,n})\in\mathbb{C}^{P/2+1}$ denote its real FFT. We represent the complex spectrum by concatenating its real and imaginary components,
\begin{equation}
    \mathbf{f}_{c,n}
    =
    \left[
        \operatorname{Re}\mathcal{F}(\mathbf{x}_{c,n});
        \operatorname{Im}\mathcal{F}(\mathbf{x}_{c,n})
    \right]
    \in\mathbb{R}^{2(P/2+1)}.
    \label{eq:rfft}
\end{equation}
The temporal and spectral views are embedded separately by learnable projections $W_t$ and $W_f$, and fused by $ W_{fuse} $

\begin{equation}
    \mathbf{h}^{t}_{c,n}=W_{t}\mathbf{x}_{c,n},
    \qquad
    \mathbf{h}^{f}_{c,n}=W_{f}\mathbf{f}_{c,n},
\end{equation}
\begin{equation}
    \mathbf{h}_{c,n}
    =W_{\mathrm{fuse}}
    \left[
        \mathbf{h}^{t}_{c,n};
        \mathbf{h}^{f}_{c,n}
    \right].
    \label{eq:fusion}
\end{equation}
where $\mathbf{h}^{t}_{c,n}\in\mathbb{R}^{d_t}$ and $\mathbf{h}^{f}_{c,n}\in\mathbb{R}^{d_f}$ are the temporal and spectral representations of patch $(c,n)$. Their concatenation is projected by $W_{\mathrm{fuse}}$ into a shared fused representation $\mathbf{h}_{c,n}\in\mathbb{R}^{d_h}$.
The fused representation is first normalized. To preserve the structure of the multivariate input, we add a two-dimensional positional encoding that identifies both the channel and temporal location of each patch. Specifically, we use a channel embedding $\mathbf{p}^{ch}_{c}$ and a patch-position embedding
$\mathbf{p}^{pos}_{n}$, giving
\begin{equation}
    \widetilde{\mathbf{h}}_{c,n}
    =
    \operatorname{LN}(\mathbf{h}_{c,n})
    + \mathbf{p}^{ch}_{c}
    + \mathbf{p}^{pos}_{n}.
    \label{eq:2d_pe}
\end{equation}
Here, LN denotes layer normalization. A transformer decoder then maps the patch representation to the language-model width $d_{\mathrm{model}}$. We write the complete Time-series Encoder module in Figure~\ref{fig:overview}(a), including the projector, as
\begin{equation}
    \mathbf{z}_{c,n}
    =g_{\phi}(\mathbf{x}_{c,n},c,n)
    \in\mathbb{R}^{d_{\mathrm{model}}},
    \label{eq:ts_encoder}
\end{equation}
where $\phi$ denotes all trainable parameters of the time-series encoder. The vectors $\mathbf{z}_{c,n}$ are the $CN$ soft time-series tokens shown in the figure.

\paragraph{Token Sequence.}
The encoded time series are inserted into the language-model input as soft tokens while preserving the original ordering of the multimodal input. Each time series \(\mathbf{S}\) is mapped by \(g_{\phi}(\mathbf{S})\) to a sequence of soft tokens and enclosed by learned \(\langle ts\rangle\) and \(\langle/ts\rangle\) delimiters. When an input contains multiple time series, each series is represented by its own delimited soft-token sequence, with any intervening text retained between them. Thus, text, captions, and time-series representations are interleaved in the same order as they appear in the original example. The target answer \(A\) is appended during teacher-forced training.
The resulting multimodal token sequence, denoted by \(\mathbf{E}\), is processed by the pretrained causal language model,
\begin{equation}
\mathbf{H}=LLM_{\Theta}(\mathbf{E}),
\label{eq:lm}
\end{equation}
where \(\mathbf{H}\) denotes the output hidden representations for all positions in the sequence.


\subsection{Phase 1: Forecast Grounding Pretraining}
\label{sec:pg}

Forecast grounding pretraining uses a mixture of forecasting, captioning, and time-series question-answering data, as shown in Figure~\ref{fig:overview}(b). The time-series encoder and projector are trained jointly with the next-patch decoder, while the language model is adapted through LoRA. We denote the encoder parameters by $\phi$, the next-patch decoder parameters by $\psi$, and the LoRA parameters by $\eta$.

\paragraph{Language Objective.}
For language generation tasks like captioning and question-answering examples, the supervision is a standard autoregressive language objective. Let $\mathcal{A}$ denote the target answer positions,  and $a_t$ denote the target token at position $t$. We optimize:
\begin{equation}
    \mathcal{L}_{\mathrm{lang}}
    =-
    \sum_{t\in\mathcal{A}}
    \log p_{\phi,\eta}
    \left(a_t\mid\mathbf{E}_{<t}\right).
    \label{eq:lang}
\end{equation}
This objective jointly updates the time-series encoder's parameters $\phi$ and the LoRA parameters $\eta$ through the language-model computation.

\paragraph{Forecasting Objective.}
Forecasting examples provide the predictive component of pretraining.
For patch $\mathbf{x}_{c,n}$, let $u(c,n)$ denote the corresponding
time series token position. The next-patch decoder $d_{\psi}$
predicts the normalized succeeding patch from the final hidden states of the LLM:
\begin{equation}
    \widehat{\mathbf{x}}_{c,n+1}
    =
    d_{\psi}\!\left(
        \mathbf{H}_{u(c,1)},\ldots,\mathbf{H}_{u(c,n)}
    \right)_{n}.
    \label{eq:next_patch}
\end{equation}
In implementation, \(d_{\psi}\) projects the language-model hidden states back to the time-series representation space and applies a causally masked Transformer decoder, followed by a linear prediction head that outputs the \(P\) values of the succeeding patch. The causal Transformer allows the prediction at position \(n\) to condition on \(\mathbf{H}_{u(c,n)}\) as well as preceding states from the same channel.

Let
$
    \mathcal{V}
    =
    \left\{
        (c,n):
        1 \leq c \leq C,\;
        1 \leq n < N
    \right\}$
denote the set of positions with an observed successor patch in the
same channel. Sequence-padding patches and positions without a
successor are excluded. The forecasting loss is
\begin{equation}
    \mathcal{L}_{\mathrm{pred}}
    =
    \frac{1}{|\mathcal{V}|}
    \sum_{(c,n)\in\mathcal{V}}
    \left\|
        \widehat{\mathbf{x}}_{c,n+1}
        -
        \mathbf{x}_{c,n+1}
    \right\|_2^2.
    \label{eq:pred}
\end{equation}

The decoder therefore operates on the LLM's hidden state. Gradients from $\mathcal{L}_{\mathrm{pred}}$ propagate through the language-model computation and jointly update the time-series encoder's parameters $\phi$, the LoRA parameters $\eta$, and the next-patch decoder's parameters $\psi$. 

\paragraph{Optimization.}
We train on separate forecasting and language batches, so each optimization step uses either $\mathcal{L}_{\mathrm{lang}}$ or $\mathcal{L}_{\mathrm{pred}}$. Let $D_{lang}$ and $D_{pred}$ denote the language and forecasting data distributions, respectively.
The overall pretraining objective is

\begin{equation}
    \mathcal{L}_{\mathrm{PG}}
    =
    \mathbb{E}_{b\sim\mathcal{D}_{\mathrm{lang}}}
    [\mathcal{L}_{\mathrm{lang}}(b)]
    +
    \mathbb{E}_{b\sim\mathcal{D}_{\mathrm{pred}}}
    [\mathcal{L}_{\mathrm{pred}}(b)].
    \label{eq:pg}
\end{equation}
Because the Phase-1 corpora differ greatly in size, size-proportional sampling would cause the largest forecasting corpus to dominate training. We therefore upsample smaller sources and downsample larger ones according to predefined mixture weights, making the language--forecasting balance independent of corpus size. Language batches update $\phi$ and $\eta$ through Eq.~\ref{eq:lang}; forecasting batches update $\phi$, $\eta$, and $\psi$ through Eq.~\ref{eq:pred}. 




\subsection{Phase 2: Downstream Adaptation}
\label{sec:downstream}
After pretraining, we remove the next-patch decoder and jointly adapt one model across downstream tasks, fully fine-tuning the encoder and projector while applying LoRA to the LLM:
\begin{equation}
\min_{\phi,\eta}
\mathbb{E}_{b\sim\mathcal{D}_{\mathrm{down}}}
[\mathcal{L}_{\mathrm{lang}}(b)].
\label{eq:downstream}
\end{equation}
We balance the five corpora by sampling $B=40{,}000$ examples from each per epoch. For a corpus of size $N_k$, we repeat the full corpus $q_k=\lfloor B/N_k\rfloor$ times, then fill the remaining $r_k=B-q_kN_k$ slots with examples selected without replacement using a seeded shuffle. When $N_k>B$, no full repetition is needed; we select a fixed subset of $B$ examples and reuse it across epochs. The resulting $200{,}000$-example mixture gives each corpus equal representation, preventing larger datasets from dominating the sample count (more details in Appendix~\ref{sec:app:phase2data}).

\section{Experiments}
\subsection{Datasets and Implementation Details}
\label{sec:exp:data}

\paragraph{Phase 1.} Our pretraining phase uses four datasets: half-hourly electricity demand from Monash~\citep{godahewa2021monash}, our synthetic multivariate corpus (details in Appendix~\ref{sec:app:phase1-data}), TSQA multiple-choice questions on trend, volatility, seasonality, and outliers~\citep{kong2025timemqa}, and M4 series with generated captions~\citep{makridakis2020m4}. We retain OpenTSLM's language pretraining data~\citep{opentslm}, adding Monash and our synthetic as forecasting pretraining datasets.


\paragraph{Phase 2.} The downstream adaptation phase finetunes the checkpoints of Phase 1 on five datasets: HAR~\citep{opentslm}, PAMAP2~\citep{reiss2012pamap2}, Sleep-EDF~\citep{kemp2000sleepedf}, ECG-QA~\citep{ecgqa,ptbxl}, and TimeMQA~\citep{kong2025timemqa}, whose MCQ and T/F splits are evaluated separately. Unless otherwise stated, we finetune our method and all baselines on those datasets jointly (as mentioned in Section \ref{sec:downstream}). Each corpus receives a budget of $40{,}000$ examples so that none dominates the mixture. Details on phase 2 datasets are in Appendix~\ref{sec:app:phase2data}.

\paragraph{Training and Evaluation Protocols.} 
For training, during Phase~1, the encoder, projector, decoder, and LLM LoRA adapters are trained jointly. During Phase~2, the next-patch decoder is removed, and the time-series interface and LLM LoRA adapters are trained using the language objective alone. Appendix~\ref{sec:app:training} provides the complete optimization and LoRA configurations. For evaluation, every baseline is evaluated under greedy decoding: the model writes a rationale and an explicit answer, which is parsed and compared with the reference. We report classification accuracy on HAR, PAMAP2, Sleep-EDF, and TimeMQA, and generative accuracy under normalization on ECG-QA. Appendix~\ref{sec:app:evaluation} details answer normalization and the held-out benchmarks.

\begin{table}[t]
\centering
\setlength{\tabcolsep}{3.5pt}
\renewcommand{\arraystretch}{0.94}
\caption{\textbf{Accuracy on downstream tasks.} Avg.\ is the mean over the six splits; \# Params counts trainable parameters in downstream adaptation. Per backbone block, \textbf{bold} is best and \underline{underline} second best per column; $\dagger$ marks OpenTSLM's original evaluation protocol~\citep{opentslm}.}
\label{tab:main}
\footnotesize
\begin{tabular}{c|lr|ccccccc}
\toprule
& Baseline & \# Params & HAR & Sleep & ECG & PAMAP2 & MCQ & T/F & \textbf{Avg.} \\
\midrule
\multirow{8}{*}{\rotatebox[origin=c]{90}{Reference}}
& Chance & --- & 12.50 & 14.29 & 0.98 & 8.33 & 25.00 & 50.00 & 18.52 \\
& Llama-3.1-8B-Instruct, zero-shot & 0 & 9.26 & 24.52 & 18.36 & 9.83 & 8.74 & 9.42 & 13.35 \\
& Qwen2.5-7B-Instruct, zero-shot & 0 & 15.97 & 31.40 & 31.76 & 9.40 & 10.02 & 3.34 & 16.98 \\
& Mistral-7B-Instruct-v0.3, zero-shot & 0 & 7.07 & 21.40 & 21.31 & 6.84 & 24.31 & 24.62 & 17.59 \\
& Gemma3-4B-pt (image), zero-shot & 0 & 1.97 & 0.00 & 0.00 & 0.00 & 17.48 & 31.61 & 8.51 \\
& Gemma3-4B-pt (image), FT & 14.9M & 51.31 & 61.18 & 30.94 & 51.71 & 46.10 & 70.85 & 52.02 \\
& TS-Reasoner-7B, zero-shot & 0 & 17.26 & 29.89 & 13.84 & 15.81 & 36.25 & 0.00 & 18.84 \\
& TS-Reasoner-7B, FT & 37.6M & 59.62 & 60.65 & 30.09 & 58.80 & 54.39 & 68.42 & 55.33 \\
\midrule
\multirow{7}{*}{\rotatebox[origin=c]{90}{Gemma3-270M}}
& Text-tokenized, zero-shot & 0 & 0.00 & 0.00 & 0.00 & 0.00 & 15.57 & 22.80 & 6.40 \\
& Text-tokenized, FT & 1.9M & \textbf{64.00} & 64.14 & 23.42 & 47.44 & 38.28 & \underline{72.81} & \underline{51.68} \\
& OpenTSLM-SP & 3.1M & 54.38 & 65.12 & 9.66 & 52.14 & \underline{38.59} & 68.39 & 48.05 \\
& \quad \seqarrow{} $\textit{sequential}^{\dagger}$ & 3.1M & 1.47 & 0.00 & 2.56 & 0.00 & 41.36 & 70.52 & 19.32 \\
& OpenTSLM-Flamingo & 317M & \underline{62.55} & \underline{68.01} & \underline{30.10} & \underline{54.70} & 12.37 & 3.04 & 38.46 \\
& \quad \seqarrow{} $\textit{sequential}^{\dagger}$ & 317M & 3.77 & 0.00 & 0.00 & 9.40 & 20.04 & 4.86 & 6.34 \\
& \lc \method{} & \lc 1.9M & \lc 58.38 & \lc \textbf{70.54} & \lc \textbf{33.16} & \lc \textbf{55.15} & \lc \textbf{39.41} & \lc \textbf{74.47} & \lc \textbf{55.19} \\
\midrule
\multirow{7}{*}{\rotatebox[origin=c]{90}{Gemma3-1B}}
& Text-tokenized, zero-shot & 0 & 0.00 & 0.00 & 3.46 & 0.00 & 5.76 & 10.35 & 3.26 \\
& Text-tokenized, FT & 6.5M & 57.08 & 70.65 & \underline{35.24} & \underline{51.71} & \underline{44.70} & \underline{78.12} & \underline{56.25} \\
& OpenTSLM-SP & 5.5M & 55.46 & \underline{72.58} & 21.66 & 47.44 & 41.36 & 77.51 & 52.67 \\
& \quad \seqarrow{} $\textit{sequential}^{\dagger}$ & 15.3M & 0.00 & 0.00 & 3.68 & 0.00 & 42.22 & 76.90 & 20.47 \\
& OpenTSLM-Flamingo & 1.3B & \underline{60.01} & 66.99 & 32.48 & \underline{51.71} & 19.62 & 6.69 & 39.58 \\
& \quad \seqarrow{} $\textit{sequential}^{\dagger}$ & 1.3B & 0.15 & 0.00 & 1.88 & 1.28 & 20.90 & 2.43 & 4.44 \\
& \lc \method{} & \lc 11.1M & \lc \textbf{62.67} & \lc \textbf{77.10} & \lc \textbf{37.66} & \lc \textbf{52.56} & \lc \textbf{44.98} & \lc \textbf{78.42} & \lc \textbf{58.90} \\
\midrule
\multirow{7}{*}{\rotatebox[origin=c]{90}{Qwen3-0.6B}}
& Text-tokenized, zero-shot & 0 & 5.86 & 1.51 & 13.24 & 4.70 & 36.03 & 16.11 & 12.91 \\
& Text-tokenized, FT & 5.0M & 51.40 & 34.84 & 29.66 & 30.34 & 52.88 & 70.52 & 44.94 \\
& OpenTSLM-SP & 27.8M & \underline{63.25} & \underline{70.54} & \textbf{32.20} & \underline{47.44} & \underline{54.64} & \underline{71.20} & \underline{56.55} \\
& \quad \seqarrow{} $\textit{sequential}^{\dagger}$ & 27.8M & 4.00 & 30.54 & 5.02 & 3.42 & 48.19 & 72.37 & 27.26 \\
& OpenTSLM-Flamingo & 650M & 60.54 & 45.27 & 28.12 & 46.58 & 37.13 & 21.88 & 39.92 \\
& \quad \seqarrow{} $\textit{sequential}^{\dagger}$ & 650M & 26.51 & 40.97 & 4.86 & 14.53 & 31.13 & 18.24 & 22.71 \\
& \lc \method{} & \lc 9.3M & \lc \textbf{63.66} & \lc \textbf{77.20} & \lc \underline{31.58} & \lc \textbf{53.42} & \lc \textbf{56.72} & \lc \textbf{76.90} & \lc \textbf{59.91} \\
\bottomrule
\end{tabular}
\end{table}



\subsection{Baselines}
\label{sec:exp:baselines}
We include a set of well-known baselines with different model sizes, backbones and input modalities in \textbf{Reference}, including fine-tuned Gemma3-4B \citep{gemma3} and fine-tuned TS-Reasoner \citep{yu2026tsreasoner}. We then match backbones and sizes across \textbf{Text-tokenized fine-tuning}, using digit strings and LoRA~\citep{gruver2023llmtime}; \textbf{OpenTSLM-SP}, using soft prompts; and \textbf{OpenTSLM-Flamingo}, using gated cross-attention~\citep{opentslm}. Each uses Gemma3-270M, Gemma3-1B \citep{gemma3}, and Qwen3-0.6B \citep{yang2025qwen3technicalreport}, sharing \method{}'s Phase~1 source corpora with method-specific supervision and Phase~2 downstream datasets. OpenTSLM rows marked with \textit{sequential$^\dagger$} follow the original sequential fine-tuning recipe~\citep{opentslm}.

\subsection{Results}
\label{sec:results}
\label{sec:results:main}


\paragraph{Main Results.} 
\begin{wrapfigure}{r}{0.45\textwidth}
    \vspace{-\intextsep}
    \centering
    \includegraphics[width=0.45\textwidth]{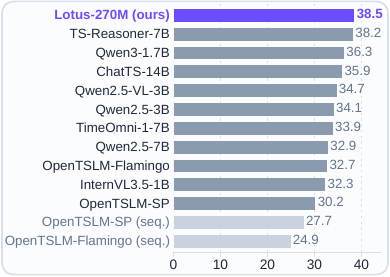}
    \caption{\textbf{TSRBench overall accuracy}. Per-subset results are in Appendix's Table~\ref{tab:tsrbench-full}.}
    \label{fig:tsrbench}
    \vspace{-\intextsep}     
\end{wrapfigure}
We present the main results in Table~\ref{tab:main}. \method{} achieves the best six-task average among the compared methods at each backbone size, and performs best in every dataset except HAR at the smallest Gemma model and ECG-QA on Qwen3. We achieve these results with far fewer trainable parameters than other baselines: on Qwen3-0.6B we train a third of what OpenTSLM-SP does and roughly a seventieth of what OpenTSLM-Flamingo does. Our smallest model, \method{} on Gemma-270M comes within $0.14$\% of a fully fine-tuned TS-Reasoner-$7$B system ($55.19\%$ vs.\ $55.33\%$) whose backbone and trainable parameters are an order of magnitude larger. Overall, \method{} on Qwen3-0.6B achieves the best average result of $59.91$\%, $4.58$\% above TS-Reasoner-7B-FT, while being \textbf{14 times smaller in backbone size}.

\paragraph{Sequential adaptation substantially degrades earlier-task performance.}
The \textit{sequential$^\dagger$} baselines in Table~\ref{tab:main} follow OpenTSLM's original sequential fine-tuning recipe and dataset order (HAR $\rightarrow$ Sleep $\rightarrow$ ECG), with PAMAP2 and TimeMQA appended. Compared with balanced joint-mixture training, sequential adaptation lowers average accuracy by $17.21\%$--$35.14\%$. Performance declines mainly on datasets encountered earlier, including eight settings with $0\%$ accuracy. In contrast, accuracy on TimeMQA's MCQ and T/F splits, trained last, is largely preserved and sometimes matches the jointly trained counterparts. This pattern suggests that adapting to later tasks can compromise performance on earlier ones. We therefore use joint adaptation for the main comparison, exposing the shared encoder and projector to all datasets throughout training.

\paragraph{Zero-shot Results On TSRBench.} 
We evaluate zero-shot reasoning ability on TSRBench~\citep{tsrbench}, using our Gemma3-270M checkpoint. As Figure~\ref{fig:tsrbench} shows, \textbf{\method{}-270M scores highest among the thirteen baselines}, ahead of \textbf{backbones 25--50$\times$ larger} like TS-Reasoner-7B and ChatTS-14B. The per-subset results in Table~\ref{tab:tsrbench-full}. \method{} achieves the highest overall TSRBench accuracy among the evaluated models, with strong performance in event prediction and comparative analysis. From the results, we believe \method{} transferable strength shows most in reasoning from signal evidence to a conclusion. Detailed per-task results are reported in Appendix~\ref{app:tsrbench}.


\section{Analysis}
\label{sec:analysis}

\subsection{Component ablation}
\label{sec:analysis:ablation}

\begin{figure}[t]
\centering
\begin{subfigure}[b]{0.42\textwidth}
  \centering
  \includegraphics[width=\linewidth]{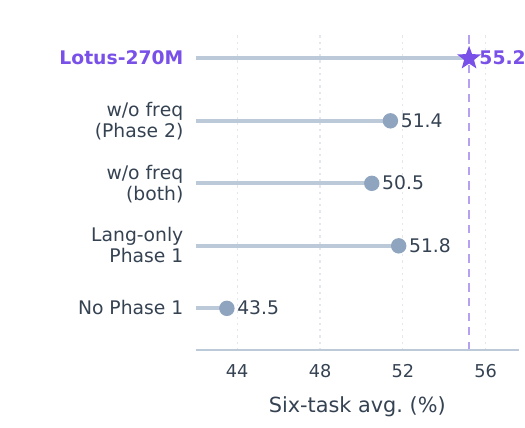}
  \caption{Component ablation.}
  \label{fig:ablation}
\end{subfigure}\hfill
\begin{subfigure}[b]{0.56\textwidth}
  \centering
  \includegraphics[width=\linewidth]{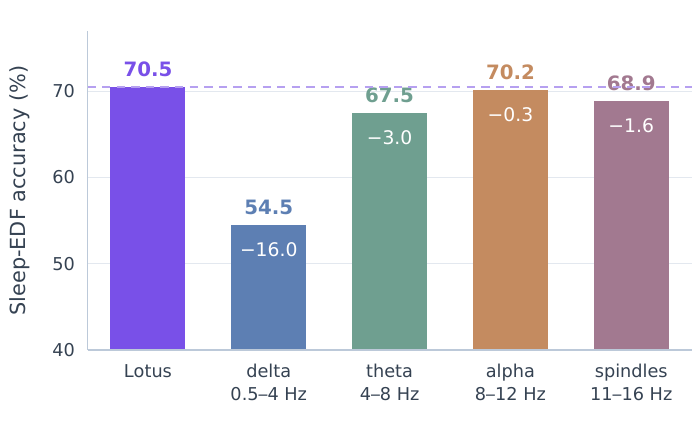}
  \caption{Frequency sensitivity.}
  \label{fig:freq}
\end{subfigure}
\caption{\textbf{Component and input analyses on Gemma3-270M.}
\textbf{(a)} Six-task average under ablations of pretraining and
the frequency branch.
\textbf{(b)} Sleep-EDF accuracy after removing individual
frequency bands from the test signal. In-bar labels show percentage-point drops.}
\label{fig:analysis}
\end{figure}

Figure~\ref{fig:ablation} compares \method{} with variants that remove pretraining or frequency information. \textit{No Phase 1} fine-tunes directly in Phase~2 with a randomly initialized encoder and projector, lowering average accuracy from $55.2\%$ to $43.5\%$. \textit{Lang-only Phase 1} uses TSQA and M4 at the same pretraining step budget and scores $3.4\%$ below full \method{} on average. Removing the frequency branch in Phase~2 alone or both phases lowers performance by $3.8\%$ and $4.7\%$ relative to full \method{}, respectively, with the largest losses on ECG-QA and PAMAP2. In these variants, each time-series patch is embedded using only the temporal projection $W_t\mathbf{x}_{c,n}$.

\subsection{Frequency Analysis}
\label{sec:analysis:frequency}

Figure~\ref{fig:freq} removes one frequency band at a time from the Sleep-EDF test signal. Suppressing delta ($0.5$--$4$\,Hz) lowers accuracy by $16.0\%$, $3.0\%$ for theta ($4$--$8$\,Hz), $1.6\%$ for spindles ($11$--$16$\,Hz), and $0.3\%$ for alpha ($8$--$12$\,Hz). The largest drop follows delta suppression, indicating that the trained model is particularly sensitive to this low-frequency content in Sleep-EDF. Appendix~\ref{app:diagnostics} provides complementary evidence of signal dependence: shuffling patches, replacing the signal with noise, or zeroing it reduces HAR accuracy to $26.9\%$, $26.5\%$ and $32.2\%$, respectively. These results show that signal information contributes to predictions beyond the accompanying text. The frequency-branch ablations examine how explicitly providing Fourier coordinates affects learning: removing the branch lowers six-task average accuracy by $3.8\%$ in Phase~2 alone and $4.7\%$ in both phases.

\subsection{Representation Analysis}
\label{sec:analysis:geometry}
 
\begin{figure}[t]
\centering
\includegraphics[width=\textwidth]{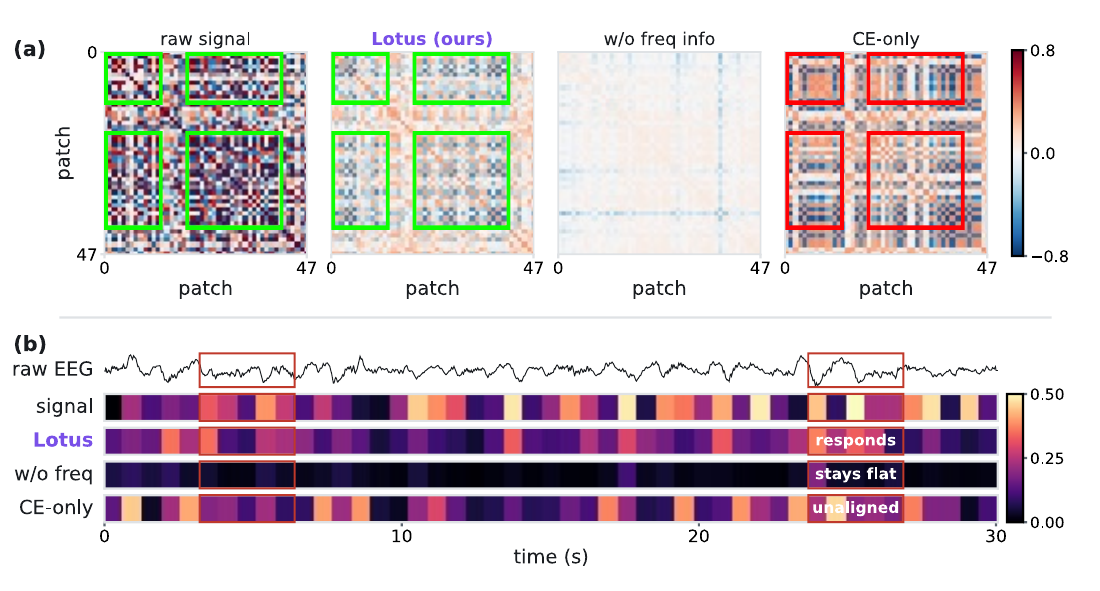}
\caption{\textbf{Representation analysis.} \textbf{(a)} Pairwise patch similarities. \textbf{\textcolor{green}{Green}} boxes highlight patterns shared by \method{} and the raw signal; \textbf{\textcolor{red}{red}} boxes mark corresponding regions where these patterns are less apparent with cross-entropy (CE)-only training. \textbf{(b)} Patch colors indicate cosine distance from the window-average representation. Boxes highlight spans selected for comparison.}

\label{fig:patch-similarity}
\end{figure}
 


We compare patch-pair similarities in raw signals and learned representations over held-out Sleep-EDF windows of $47$ patches. For patch vectors $u_i$ and $u_j$, we compute cosine similarity as $s_{ij}=u_i^\top u_j/(\|u_i\|_2\|u_j\|_2)$ and centre each panel on its off-diagonal mean. \textbf{Figure~\ref{fig:patch-similarity}a} uses patch indices on both axes, with blocks identifying groups of similar patches. \method{} (second panel) reproduces prominent raw-signal blocks, highlighted by \textcolor{green}{\textbf{green}} boxes, suggesting that patches with similar signal content retain similar representations. Without the frequency branch, the third panel shows a more uniform pattern with less visible correspondence to these structures. The cross-entropy (CE)-only representation retains varied patterns (fourth panel), but those marked by \textcolor{red}{\textbf{red}} boxes align less closely with the raw signal. \textbf{Figure~\ref{fig:patch-similarity}b} shows a Sleep-EDF EEG signal and patch-wise cosine distances $d_i=1-u_i^\top\bar{u}/(\|u_i\|_2\|\bar{u}\|_2)$ over time in seconds, where $\bar{u}=\frac{1}{N}\sum_{j=1}^{N}u_j$ is the window-average representation. Brighter colours indicate greater deviation from the window average; the raw-signal row is separately rescaled. In the highlighted spans (red boxes) for comparison, \method{} varies alongside the signal, whereas the no-frequency variant remains relatively flat and CE-only changes align less closely. Together, these findings suggest that \method{} preserves both relationships between patches and local variation within the signal.

\section{Conclusion}
\label{sec:conclusion}

We introduce LOTUS, a time-series language model built on \emph{forecast grounding}: pretraining language-aligned representations to retain information about future signal evolution. LOTUS combines next-patch prediction from LLM hidden states with a temporal--spectral patch representation. Across three language-model backbones and six downstream splits, LOTUS achieves the best average performance among matched baselines. Ablations show gains from forecast grounding and frequency information, while representation analyses suggest that predictive supervision helps preserve temporal structure. These findings support forecasting as a pretraining objective for time-series language models, even when downstream tasks require only language outputs.

\subsection*{AI Use Statement}
Large language models were not involved in the design of our approach. We used them solely to polish the writing of this paper (grammar and style). All methods, experiments, and analyses were designed and run by the authors, who take responsibility for the content.

\subsection*{Ethics Statement}
This work uses existing time-series datasets and synthetic data. Physiological and activity signals raise privacy concerns, and model predictions may reflect dataset biases or produce misleading explanations. LOTUS is evaluated as a research system; its results do not establish suitability for clinical decision-making or other high-stakes applications.

\subsection*{Reproducibility Statement}
We describe the architecture and training objectives in Section \ref{sec:method} and provide training configurations and evaluation procedures in Appendix \ref{app:experiment}. Appendices \ref{sec:app:phase1-data} and \ref{sec:app:phase2data} document data construction, splits, and sampling, while Appendices \ref{app:compute} and \ref{sec:app:prompts} report computational resources and prompt templates to support reproduction of our experiments.

\bibliography{references}
\bibliographystyle{iclr2027_conference}

\newpage
\appendix
\startcontents[appendix]
\printcontents[appendix]{}{1}{\setcounter{tocdepth}{2}\section*{Appendix Contents For \\
\textit{Lotus: Learning to Understand Time-series by Learning to Forecast}}}

\section{Experiment Details}
\label{app:experiment}
\subsection{Training Configuration}
\label{sec:app:training}

Table~\ref{tab:recipe} summarizes the recipe shared across the reported runs.
Phase~1 uses source-homogeneous batches, so each update applies either the
language or prediction loss. Phase~2 draws the same fixed example budget from
each corpus and applies only the language loss.

\subsection{Evaluation Details}
\label{sec:app:evaluation}

All matched systems use greedy decoding. Each response contains a rationale followed by an explicit answer field. The parser extracts that field, applies case folding and whitespace normalization, and maps dataset-specific textual variants to the task's canonical label. An output is counted as incorrect when no label can be recovered. ECG-QA uses the released answer normalization. The headline metric is generative accuracy, followed by an unweighted mean across the four non-overlapping datasets; it is not a token-level language-model metric.

\subsection{Training Details}
\label{sec:app:details}


We provide the detailed configs for both phases in Table~\ref{tab:recipe}.

\begin{table}[h]
\caption{Configuration of the two phases. Settings shared by both phases are listed first.}
\label{tab:recipe}
\centering
\small
\setlength{\tabcolsep}{4pt}
\begin{tabular}{@{}l p{0.33\textwidth} p{0.33\textwidth}@{}}
\toprule
 & Phase 1 & Phase 2 \\
\midrule
\rowcolor{sharedrow} \multicolumn{3}{@{}l}{\textit{Shared}} \\
Patch size & \multicolumn{2}{@{}p{0.68\textwidth}@{}}{$P = 32$, non-overlapping} \\
Encoder & \multicolumn{2}{@{}p{0.68\textwidth}@{}}{Transformer decoder, width $64$, FF $256$, output $768$; $95{,}680$ params} \\
Position encoding  & \multicolumn{2}{@{}p{0.68\textwidth}@{}}{fixed 2-D sinusoidal over (patch index, channel index); not learned} \\
Next-patch Decoder & \multicolumn{2}{@{}p{0.68\textwidth}@{}}{linear projection to the encoder width, causally masked Transformer de-
coder, linear head to Poutputs; } \\
Optimiser & \multicolumn{2}{@{}p{0.68\textwidth}@{}}{AdamW~\citep{loshchilov2019adamw}, weight decay $10^{-2}$} \\
LR (enc./proj.) & \multicolumn{2}{@{}p{0.68\textwidth}@{}}{$1\times10^{-4}$} \\
LR schedule & \multicolumn{2}{@{}p{0.68\textwidth}@{}}{linear warmup over the first $3\%$ of steps, then linear decay to $0$} \\
Grad.\ clipping & \multicolumn{2}{@{}p{0.68\textwidth}@{}}{global norm $1.0$} \\
Precision & \multicolumn{2}{@{}p{0.68\textwidth}@{}}{AMP fp16 with \texttt{GradScaler}} \\
Hardware & \multicolumn{2}{@{}p{0.68\textwidth}@{}}{$8\times$ NVIDIA V100 ($2$ nodes $\times\,4$); effective batch $128$} \\
\midrule
\rowcolor{phaserow} \multicolumn{3}{@{}l}{\textit{Phase-specific}} \\
Data & forecasting, QA, captioning & six CoT corpora \\
Mixture & weighted $1{:}2{:}1{:}1$ ($60\%$ num.\ / $40\%$ ling.) & balanced, $40{,}000$ examples per task per epoch \\
Epoch size & task-weighted sampler & $5\times40{,}000 = 200{,}000$ \\
Objective & $\mathcal{L}_{\text{CE}} + \mathcal{L}_{\text{MSE}}$ & $\mathcal{L}_{\text{CE}}$ only \\
LoRA & $r{=}8$, $\alpha{=}16$, all attention and MLP projections & $r{=}8$, $\alpha{=}16$, all attention and MLP projections (excluding Decoder's parameters) \\
LR (LoRA) & $2 \times 10^{-4}$ & $2\times10^{-4}$ \\
Batch size & $16$ & $1$ \\
Grad.\ accumulation & $1$ & $16$ \\
Selection & validation loss & mean generative EM, $192$ validation items per task \\
Early stopping & patience 5 & per task, patience $8$ \\
\bottomrule
\end{tabular}
\end{table}

\subsection{Full Training Data Ablation Results}
\label{app:ablation}
In this section, we provide the detailed results for removing elements from our training data. Table~\ref{tab:ablation} gives the per-task results behind Figure~\ref{fig:ablation}. Language-only pretraining lowers every signal task (HAR $-4.0$, Sleep $-4.1$, ECG-QA $-2.3$, PAMAP2 $-6.4$) while leaving MCQ unchanged. Removing the frequency branch is felt most on ECG-QA ($-7.4$ in Phase~2 only, $-7.9$ in both phases) and PAMAP2 ($-13.7$/$-10.7$); HAR is the one task that improves without it ($+3.2$ in the Phase-2-only variant).

\begin{table}[h]
\centering
\caption{\textbf{Ablations on Gemma3-270M, per task.} Each variant changes one ingredient and is otherwise trained and evaluated as in Table~\ref{tab:main}. \textbf{Bold} is best per column.}
\label{tab:ablation}
\footnotesize
\setlength{\tabcolsep}{4pt}
\begin{tabular}{lccccccc}
\toprule
Variant & HAR & Sleep & ECG & PAMAP2 & MCQ & T/F & Avg. \\
\midrule
\lc \method{}                        & \lc 58.38 & \lc \textbf{70.54} & \lc \textbf{33.16} & \lc \textbf{55.15} & \lc \textbf{39.41} & \lc \textbf{74.47} & \lc \textbf{55.19} \\
w/o frequency branch (Phase 2 only)   & \textbf{61.55} & 69.46 & 25.80 & 41.45 & 37.10 & 72.95 & 51.39 \\
w/o frequency branch (both phases)    & 55.89 & 67.54 & 25.24 & 44.44 & 37.31 & 72.64 & 50.51 \\
Language-only Phase 1 (same budget)   & 54.38 & 66.45 & 30.84 & 48.72 & 39.23 & 71.16 & 51.79 \\
\bottomrule
\end{tabular}
\end{table}

\section{Input Diagnostics}
\label{app:diagnostics}

\begin{figure}[h]
\centering
\includegraphics[width=0.7\textwidth]{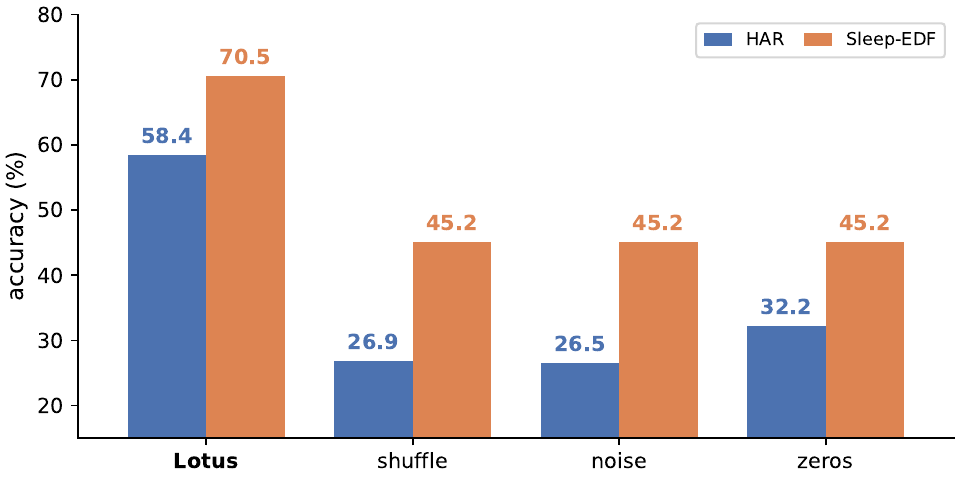}
\caption{\textbf{Input diagnostics.} Accuracy after corrupting the numerical input, on the
Gemma3-270M checkpoint and the full test splits (HAR $n{=}8{,}222$, Sleep-EDF $n{=}930$).}
\label{fig:diagnostics}
\end{figure}

We test whether predictions can be recovered from the text prompt or label prior alone. Figure~\ref{fig:diagnostics} evaluates the Gemma3-270M \method{} checkpoint of Table~\ref{tab:main} on the full test splits ($8{,}222$ HAR and $930$ Sleep-EDF items) after shuffling the signal, replacing it with Gaussian noise, or setting it to zero. HAR accuracy falls from $58.4\%$ to $26.5$--$32.2\%$, bracketing its empirical majority-class rate of $28.3\%$; Sleep-EDF falls from $70.5\%$ to $45.2\%$ under every corruption, its majority-class rate being $45.3\%$. Thus, the model's predictions depend materially on the numerical input.

\section{Per-task TSRBench Results}
\label{app:tsrbench}

Figure~\ref{fig:tsrbench} reports the aggregate; the full
per-subset breakdown behind it is given in Table~\ref{tab:tsrbench-full}.

\paragraph{Evaluation protocol.}
We follow the benchmark's evaluation~\citep{tsrbench} and use its released scorer unmodified: it extracts the option letter from the generated text, drops responses from which no letter can be recovered, and micro-averages accuracy over subsets. 

\begin{table*}[t]
\centering
\setlength{\tabcolsep}{1.5pt}
\caption{\textbf{Per-task TSRBench accuracy (\%) over 14 subsets}. Columns are grouped into perception, reasoning, prediction, and decision-making.}
\label{tab:tsrbench-full}
\scriptsize
\resizebox{\textwidth}{!}{%
\begin{tabular}{l cccc cccccc cc cc c}
\toprule
& \multicolumn{4}{c}{Perception} & \multicolumn{6}{c}{Reasoning} &
\multicolumn{2}{c}{Prediction} & \multicolumn{2}{c}{Decision} & \\
\cmidrule(lr){2-5}\cmidrule(lr){6-11}\cmidrule(lr){12-13}\cmidrule(lr){14-15}
Model & PR & NU & AD & CA & ER & CD & AR & TR & NR & IR & TSF & EP &
QualDM & QuantDM & Overall \\
\midrule
Qwen2.5-7B & 40.3 & 50.0 & 43.3 & 64.0 & 17.7 & 28.2 & 32.0 & 20.0 & 32.1 & 34.3 & 31.8 & 42.0 & 32.6 & 19.4 & 32.9 \\
Qwen3-1.7B & 39.1 & 51.2 & 33.9 & 48.5 & 16.0 & 35.3 & 42.7 & 28.1 & 28.0 & 32.0 & 40.5 & 69.4 & 32.5 & 17.7 & 36.3 \\
Qwen2.5-3B & 40.7 & 40.7 & 33.1 & 46.9 & 25.4 & 20.1 & 24.0 & 28.0 & 33.3 & 20.2 & 32.4 & 66.9 & 34.0 & 20.0 & 34.1 \\
Qwen2.5-VL-3B & 56.1 & 56.0 & 45.0 & 53.1 & 28.0 & 17.1 & 26.7 & 28.8 & 30.0 & 32.0 & 26.5 & 66.7 & 33.4 & 23.7 & 34.7 \\
InternVL3.5-1B & 47.0 & 40.2 & 34.9 & 48.2 & 23.1 & 27.7 & 24.7 & 27.5 & 27.0 & 25.0 & 39.6 & 27.6 & 32.2 & 20.3 & 32.3 \\
TimeOmni-1-7B & 42.8 & 49.3 & 37.3 & 53.6 & 24.5 & 26.1 & 32.0 & 21.0 & 21.7 & 28.8 & 30.0 & 74.7 & 28.2 & 22.7 & 33.9 \\
TS-Reasoner-7B & 48.6 & 52.9 & 38.9 & 57.1 & 22.3 & 25.0 & 54.0 & 23.9 & 36.5 & 22.4 & 28.1 & 83.8 & 33.1 & 23.3 & 38.2 \\
ChatTS-14B & 46.6 & 48.3 & 41.4 & 57.7 & 24.3 & 37.0 & 57.3 & 21.9 & 42.1 & 33.0 & 29.8 & 25.8 & 27.6 & 43.7 & 35.9 \\
\midrule
OpenTSLM-SP & 41.5 & 43.7 & 38.0 & 42.5 & 28.1 & 27.8 & 32.0 & 27.2 & 28.8 & 27.0 & 21.0 & 49.0 & 23.6 & 18.0 & 30.2 \\
\quad \seqarrow{} $\textit{sequential}^{\dagger}$ & 40.5 & 48.3 & 34.4 & 48.2 & 29.5 & 25.2 & 26.7 & 26.6 & 26.0 & 31.0 & 26.0 & 16.1 & 24.8 & 19.0 & 27.7 \\
OpenTSLM-Flamingo & 36.6 & 30.8 & 33.3 & 44.0 & 27.4 & 36.4 & 25.6 & 15.4 & 36.4 & 25.6 & 23.9 & 45.4 & 25.7 & 19.1 & 32.7 \\
\quad \seqarrow{} $\textit{sequential}^{\dagger}$ & 29.6 & 35.1 & 30.0 & 24.5 & 32.4 & 21.1 & 22.7 & 12.0 & 22.1 & 29.4 & 16.7 & 34.2 & 22.9 & 12.4 & 24.9 \\
\lc \method & \lc 26.2 & \lc 50.0 & \lc 29.0 & \lc 60.9 & \lc 27.6 & \lc 25.7 & \lc 25.0 & \lc 19.2 & \lc 34.2 & \lc 29.9 & \lc 17.6 & \lc 83.5 & \lc 25.8 & \lc 23.7 & \lc 38.5 \\
\bottomrule
\end{tabular}}
\end{table*}

\section{Qualitative Comparisons}
\begin{figure}[p]
\centering
\includegraphics[width=\textwidth]{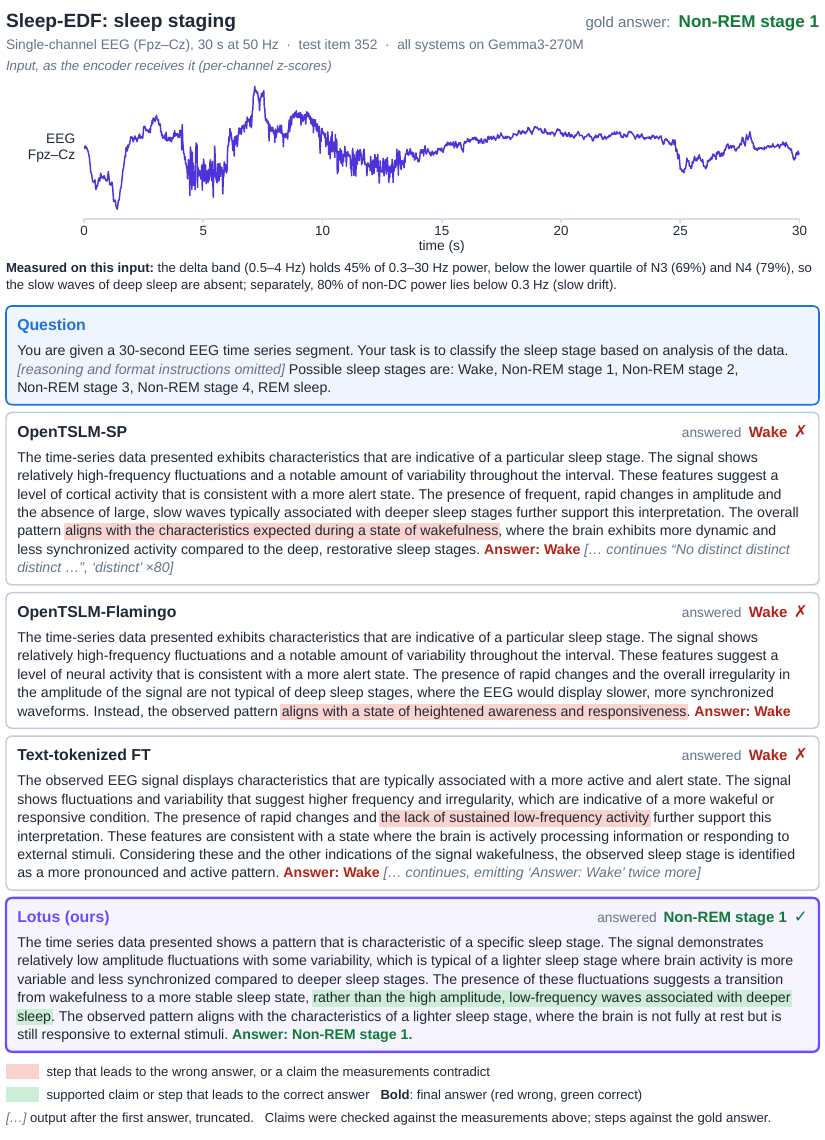}
\caption{\textbf{Qualitative example, Sleep-EDF}. All three baselines read the epoch as wakefulness; \method{} places it in a lighter sleep stage, consistent with its spectral centroid.}
\label{fig:qual-sleep}
\end{figure}

\begin{figure}[p]
\centering
\includegraphics[width=\textwidth]{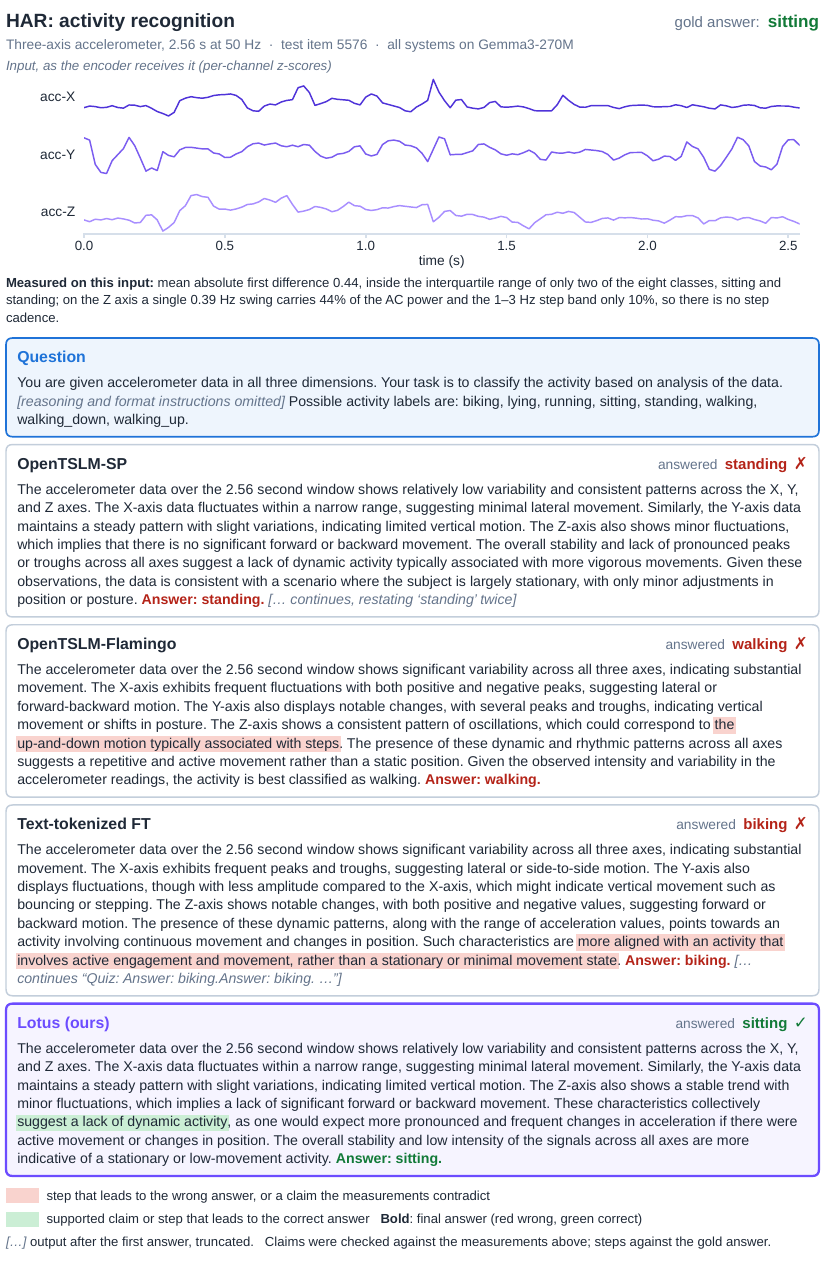}
\caption{\textbf{Qualitative example, HAR}. \method{} and OpenTSLM-SP share their first 360 characters verbatim and differ only in the final label; OpenTSLM-Flamingo describes a stepping motion that the Z-axis spectrum does not show.}
\label{fig:qual-har}
\end{figure}

\begin{figure}[p]
\centering
\includegraphics[width=\textwidth]{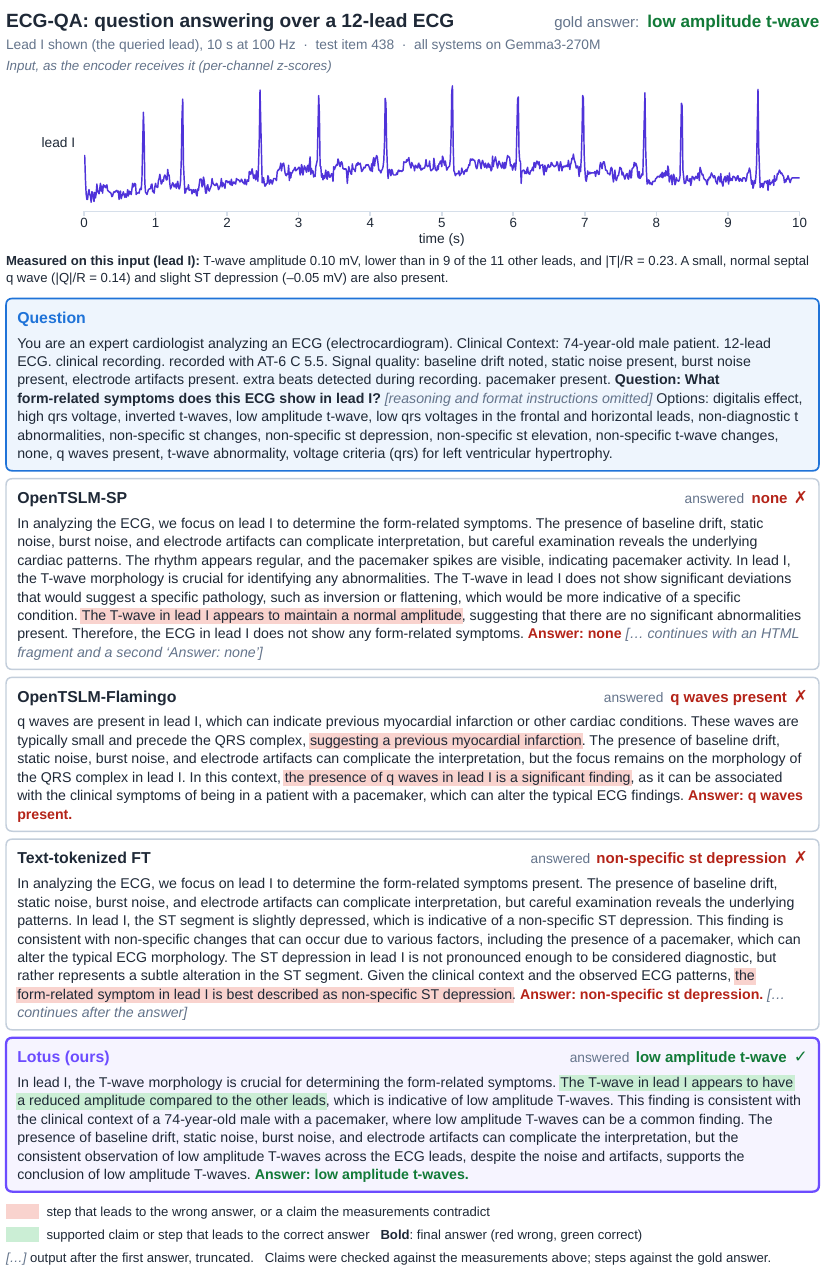}
\caption{\textbf{Qualitative example, ECG-QA}. The q wave and ST depression cited by two baselines are present in the recording but are not the queried finding; \method{} identifies the reduced T-wave amplitude in lead~I.}
\label{fig:qual-ecg}
\end{figure}

\begin{figure}[p]
\centering
\includegraphics[width=\textwidth]{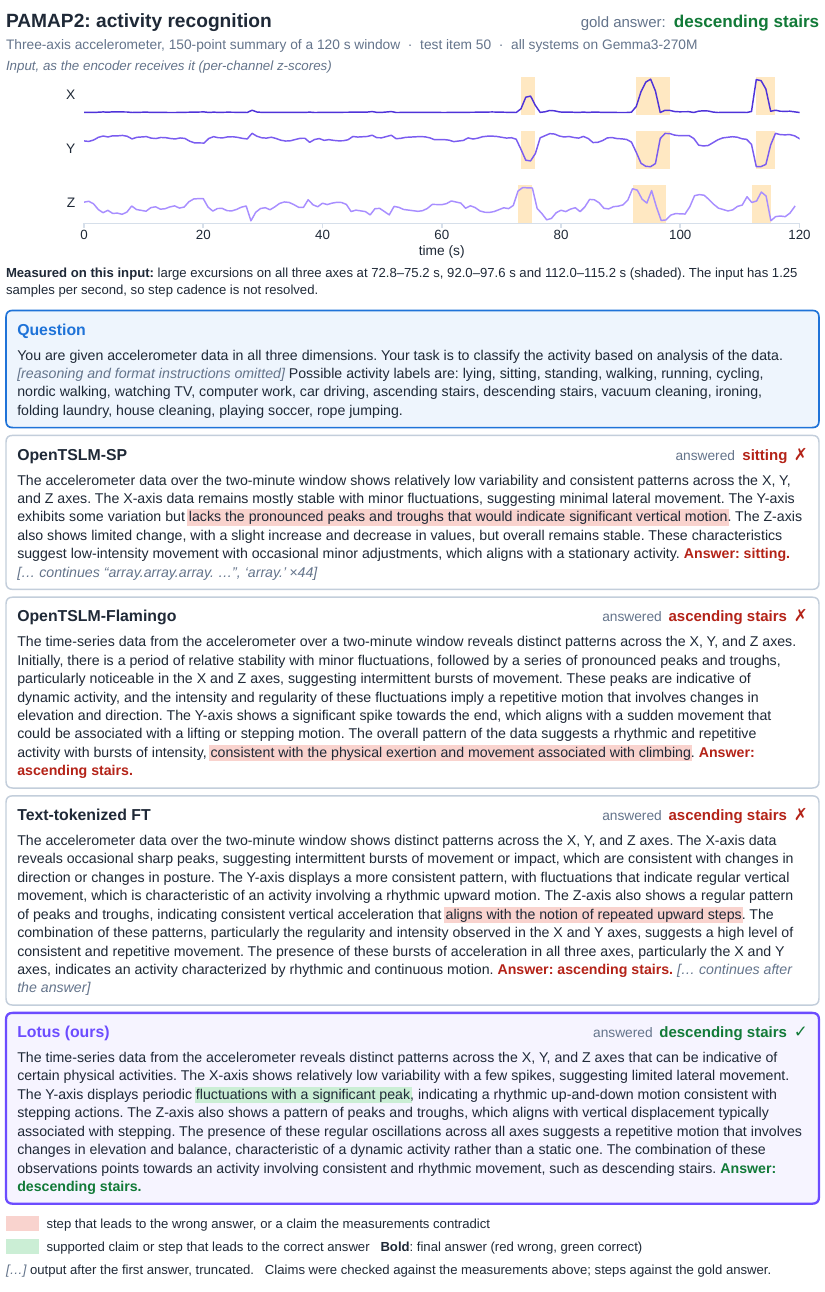}
\caption{\textbf{Qualitative example, PAMAP2}. OpenTSLM-SP misses the three shaded excursions and answers sitting. Per-channel normalization removes the gravity reference, so ascent and descent cannot be distinguished from the input; the example illustrates the missed motion rather than the direction.}
\label{fig:qual-pamap2}
\end{figure}

In this section, we provide qualitative comparisons between \method{} and the baselines on Sleep-EDF, HAR, ECG-QA, and PAMAP2. Figures~\ref{fig:qual-sleep}--\ref{fig:qual-pamap2} each show one test item with every system built on Gemma3-270M. Each figure shows the input series as the encoder receives it, the question (instructions omitted for clarity; see Appendix~\ref{sec:app:prompts}), and each model's rationale up to its first answer; any text generated after that answer is truncated. Highlights were checked against measurements on the input, reported under each series: red marks a wrong answer or a claim the measurements contradict, and green marks the correct answer and the claim it rests on. Claims the measurements support are left unmarked, including in wrong answers.

The examples show two recurring failure modes. Some baselines describe the signal correctly but answer a different question, as on ECG-QA, where two baselines report a q wave or ST depression that is present in the recording instead of the queried T-wave finding. Others misread the signal itself, as when OpenTSLM-SP describes the PAMAP2 window as stationary despite three large excursions. In all four examples, OpenTSLM-SP and the tokenized baseline also keep generating after their first answer, repeating tokens or restating the answer.

\section{Phase 1 Training Data}
\label{sec:app:phase1-data}
\subsection{Overview}

Phase~1 draws source-homogeneous batches from four corpora with fixed sampling weights $1{:}2{:}1{:}1$ (Section \ref{sec:exp:data}). The two forecasting sources therefore account for $60\%$ of optimisation steps and the two language sources for the remaining $40\%$. These proportions are determined by the sampling weights rather than by corpus size.

Two corpora provide next-patch supervision. The first contains half-hourly electricity demand for five Australian states from the Monash archive~\citep{godahewa2021monash}. We split each series chronologically into the first $80\%$ for training, the next $10\%$ for validation, and the final $10\%$ for testing. The second is the synthetic corpus described below. Because each row is generated independently rather than extracted from a shared history, we randomly partition complete rows in the same $80/10/10$ ratio; the same generated series never appears in more than one split. The two remaining corpora provide language supervision: $38.4$K multiple-choice items concerning trend, volatility, seasonality, and outliers, and $80$K M4 series~\citep{makridakis2020m4} paired with generated captions.

\subsection{Synthetic Multivariate Corpus}
\label{app:phase1-data:synth}

To obtain large-scale multivariate supervision without training on the downstream evaluation series, we construct a synthetic corpus of $844{,}275$ series. It combines four generator families (Table~\ref{tab:synth}), none of which contributes more than $37\%$ of the data. Two families impose cross-channel structure explicitly, KernelSynth contributes univariate diversity, and SymTime contributes series generated from symbolic processes.

\begin{table}[t]
\centering
\caption{Composition of the synthetic training corpus. Variate and length
ranges are observed after collection and filtering.}
\label{tab:synth}
\setlength{\tabcolsep}{6pt}
\begin{tabular}{@{}lrrcc@{}}
\toprule
Generator family & Series & Share & Variates & Length \\
\midrule
Chronos-2 multivariatizer & $310{,}000$ & $36.7\%$ & $3$--$10$ & $512$--$8{,}192$ \\
KernelSynth (univariate)  & $299{,}999$ & $35.5\%$ & $1$       & $1{,}024$--$4{,}096$ \\
TimePFN LMC               & $215{,}000$ & $25.5\%$ & $3$--$10$ & $512$--$2{,}048$ \\
SymTime S2Generator       & $19{,}276$  & $2.3\%$  & $3$--$10$ & $1{,}024$--$4{,}096$ \\
\midrule
Total                     & $844{,}275$ & $100\%$  & $1$--$10$ & $512$--$8{,}192$ \\
\bottomrule
\end{tabular}
\end{table}

\paragraph{KernelSynth.}
The univariate generator introduced by Chronos~\citep{ansari2024chronos} randomly composes a Gaussian-process (GP) covariance. For each series, it draws $j \sim \mathcal{U}\{1,\dots,5\}$ kernels with replacement from a bank of $33$ configurations. The bank contains $21$ \texttt{ExpSineSquared} configurations spanning several seasonal scales, together with parameterised \texttt{DotProduct}, \texttt{RBF}, \texttt{RationalQuadratic}, and \texttt{WhiteKernel} entries and one \texttt{ConstantKernel}. The sampled kernels are recursively combined using addition or multiplication with equal probability, and the series is drawn from the resulting GP prior on a fixed grid. Additive compositions superpose components, whereas multiplicative compositions allow one component to modulate another.

\paragraph{Linear Model of Coregionalization.}
Independent KernelSynth draws do not provide cross-channel dependence. The linear model of coregionalization (LMC) generator from TimePFN~\citep{taga2025timepfn} instead expresses each channel as a mixture of shared latent processes. For a series with $D \sim \mathcal{U}\{3,\dots,10\}$ channels, we draw $w$ from a Weibull distribution with shape $2$ and set $Q$ to $\operatorname{round}(10w+1)$, clipped to $[\max(2,\lfloor D/20\rfloor),D]$. Each of the $Q$ latents receives an independently composed KernelSynth covariance and is sampled from the corresponding GP prior. The observed channels are convex mixtures $Z=WY$, with each row of $W\in\mathbb{R}^{D\times Q}$ drawn from $\operatorname{Dir}(\alpha\mathbf{1}_Q)$ and $\alpha\sim\mathcal{U}(0.1,2.0)$. Small values of $\alpha$ produce sparse mixtures dominated by a few latents, whereas larger values distribute weight more evenly across the shared latents. Sampling $\alpha$ independently for each series yields a range of cross-channel dependence patterns.

\paragraph{Chronos-2 Multivariatizer.}
The largest family follows the synthetic-data design described in the Chronos-2 technical report~\citep{chronos2}; because its implementation is not released, we reimplement the procedure from the report. Each latent univariate series is drawn uniformly from four generators: KernelSynth; a stationary AR($p$) process with $p\sim\mathcal{U}\{1,\dots,5\}$, constructed by mapping partial autocorrelations in $(-0.95,0.95)$ to AR coefficients through the Levinson recursion; a trend--seasonality--irregularity model combining a random polynomial trend, one to three sinusoids, and Gaussian noise; or an additive Holt--Winters state-space simulator with random smoothing constants and seasonal period. After standardising the latent series, we combine $K$ of them using one of two multivariatizers, selected with equal probability. The \emph{cotemporaneous} multivariatizer computes $Z=f(AY)$, where $A\in\mathbb{R}^{D\times K}$ has i.i.d.\ $\mathcal{N}(0,1/K)$ entries. With probability $0.5$, $f$ is the identity; otherwise it is sampled uniformly from $\{\tanh,\sin,\operatorname{sgn}(x)\sqrt{|x|}\}$. This branch produces either linear or nonlinear instantaneous dependence within a series. The \emph{sequential} multivariatizer instead computes $Z[d,t]=\sum_k a_{d,k}Y_k[t-\tau_{d,k}]$, with independently sampled lags $\tau_{d,k}\sim\mathcal{U}\{0,\dots,32\}$ and zero padding at the leading boundary, thereby introducing lead--lag relationships. Mixing shared non-stationary latents in the cotemporaneous branch can also produce cointegrated channels; we do not use a separate cointegration generator. Finally, with probability $0.3$, between one and $\lfloor D/2\rfloor$ channels are designated as known covariates, yielding a covariate-informed forecasting
item.

\paragraph{SymTime.}
SymTime~\citep{wang2025synthetic} provides a qualitatively different source of variation through its series--symbol generator, which samples a symbolic expression and then realises the corresponding series. Although the generator emits matched expression--series pairs, we retain only the series because Phase~1 applies next-patch rather than symbolic-recovery supervision.

\paragraph{Unification.}
The four generators emit different container formats. During collection, we cast each item to a \texttt{float32} array with shape \texttt{(variates, length)} and stream it to a unified Arrow shard. We reject items containing non-finite values; this filter removes $1{,}668$ SymTime items and no items from the other families. The retained corpus occupies $30$\,GB in $16$ Arrow files. At training time, all files share one corpus label so that the sampler treats them as a single mixture component. Per-item seeds make the generated corpora reproducible under the same software environment.

\subsection{Dataset Statistics}
Table~\ref{tab:synth-statistic} summarises the corpus with five scale-invariant descriptors, alongside the real forecasting corpus used in the same phase. Two things are visible. First, the families occupy different regions rather than being four parameterisations of one process: KernelSynth has the lowest spectral entropy, its composed kernels concentrating spectral power and leaving the series comparatively predictable, whereas SymTime has the highest together with by far the weakest periodicity, its symbolic realisations lying closer to noise. Second, the two multivariatizers impose different amounts of cross-channel dependence, the coregionalization mixture producing the more strongly coupled channels, which is what those mechanisms are for.

The real corpus falls inside the synthetic range on spectral entropy and trend, but outside it on both dependence measures: its half-hourly demand is more persistent than any synthetic family, and its five states are more tightly coupled than any generator produces. The synthetic corpus is, on the other hand, much more varied, its per-descriptor spread being an order of magnitude wider, since the real reference is a single system observed repeatedly whereas each synthetic draw resamples its own generating process. Pretraining on this mixture therefore buys breadth rather than fidelity to any one real corpus.

\begin{table}[t]

\centering
\caption{\textbf{Phase 1 dataset statistics}, by generator family. Mean $\pm$ standard deviation over sampled series; real series are cut into $2{,}048$-point windows. Spectral entropy is spectral flatness ($0$ a pure tone, $1$ white noise); seasonal peak the largest $|\text{ACF}|$ over lags $[2,L/4]$; trend $R^2$ the variance share of a least-squares line; cross-channel the mean $|$correlation$|$ over variate pairs, undefined when univariate.}
\small
\setlength{\tabcolsep}{4pt}
\label{tab:synth-statistic}
\begin{tabular}{@{}lrccccc@{}}
\toprule
& & Spectral & & Seasonal & Trend & Cross- \\
Source & $n$ & entropy & ACF$(1)$ & peak & $R^{2}$ & channel \\
\midrule
Chronos-2 multivariatizer & $1{,}204$ & $0.51{\pm}0.14$ & $0.63{\pm}0.27$ & $0.78{\pm}0.14$ & $0.22{\pm}0.16$ & $0.36{\pm}0.14$ \\
KernelSynth (univariate)  & $1{,}206$ & $0.40{\pm}0.22$ & $0.77{\pm}0.34$ & $0.85{\pm}0.27$ & $0.10{\pm}0.24$ & -- \\
TimePFN LMC               & $1{,}202$ & $0.45{\pm}0.14$ & $0.66{\pm}0.28$ & $0.80{\pm}0.16$ & $0.07{\pm}0.12$ & $0.59{\pm}0.20$ \\
SymTime S2Generator       & $1{,}204$ & $0.75{\pm}0.16$ & $0.37{\pm}0.22$ & $0.41{\pm}0.21$ & $0.10{\pm}0.13$ & $0.24{\pm}0.19$ \\
\midrule
Australian electricity (real) & $1{,}200$ & $0.40{\pm}0.02$ & $0.98{\pm}0.00$ & $0.93{\pm}0.02$ & $0.03{\pm}0.02$ & $0.80{\pm}0.07$ \\
\bottomrule
\end{tabular}
\end{table}

\section{Phase-2 Training Data}
\label{sec:app:phase2data}

Phase~2 uses the five chain-of-thought corpora in Table~\ref{tab:phase2data}, each rendered as question--rationale--answer triples: HAR (three-axis accelerometry), Sleep-EDF (single-channel EEG), ECG-QA (twelve-lead ECG), PAMAP2 (household and exercise activities from wearable accelerometry), and TimeMQA (open-ended questions over univariate series). TimeMQA is one training corpus but two evaluation columns, separated by its \texttt{question\_format} field, which is why five corpora yield the six columns of Table~\ref{tab:main}.

\begin{table}[t]
\centering
\setlength{\tabcolsep}{5pt}
\caption{\textbf{Phase-2 corpora.} Natural split sizes and the balanced per-epoch count; multiplicity is how often an example is seen per epoch, with corpora larger than the target truncated to a fixed subset.}
\label{tab:phase2data}
\footnotesize
\begin{tabular}{lrrrrl}
\toprule
Corpus & Train & Val & Test & \# Train Per Epoch & Multiplicity \\
\midrule
HAR        & $68{,}542$  & $8{,}718$  & $8{,}222$  & $40{,}000$ & $0.58\times$ (fixed subset) \\
Sleep-EDF  & $7{,}434$   & $930$      & $930$      & $40{,}000$ & $5$--$6\times$ \\
ECG-QA     & $159{,}313$ & $31{,}137$ & $41{,}093$ & $40{,}000$ & $0.25\times$ (fixed subset) \\
PAMAP2     & $4{,}207$   & $234$      & $234$      & $40{,}000$ & $9$--$10\times$ \\
TimeMQA    & $33{,}764$  & $1{,}876$  & $1{,}876$  & $40{,}000$ & $1$--$2\times$ \\
\midrule
Total      & $273{,}260$ &            &            & $200{,}000$ & \\
\bottomrule
\end{tabular}
\end{table}

\subsection{Balanced Mixture.} Each corpus is resized to $40{,}000$ examples per epoch ($200{,}000$ in total): smaller corpora are repeated in full with a seeded shuffled remainder, larger ones are shuffled once and truncated to a subset that is reused in every epoch. Under the natural sizes, ECG-QA and HAR would supply $83\%$ of the data and PAMAP2 $1.5\%$; after balancing, each PAMAP2 example is seen nine to ten times per epoch while three quarters of ECG-QA is never used.

\subsection{Evaluation Subsets.} Sleep-EDF and PAMAP2 have validation and test splits of equal size ($930$ and $234$). For ECG-QA we score the first $5{,}000$ test items, pinned by a manifest keyed on a hash of the gold answer so that every system is evaluated on the same subset.

\section{Computational Cost}
\label{app:cost}

How much of the backbone's context a series occupies decides both what it costs to read that series at inference and what it costs to train on it. Everything in this section is measured on one GPU and one backbone (Gemma3-270m) so that the baselines can be compared directly; the resources actually consumed by our training runs are reported in Appendix~\ref{app:compute}.

\subsection{Reading A Series}
\begin{figure}[t]
\centering
\includegraphics[width=0.9\textwidth]{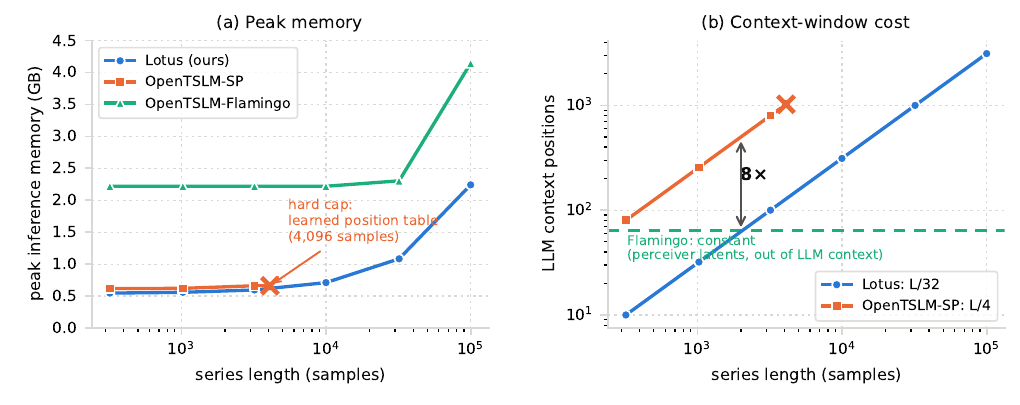}
\caption{\textbf{Inference scaling with series length}, all baselines on Gemma3-270m (single series, one $10$-token generation per point; peak allocated memory). Left: \method{} is cheapest at every length; OpenTSLM-SP's line terminates at its architectural length ceiling. Right: context positions consumed, $L/32$ against $L/4$.}
\label{fig:memory-scaling}
\end{figure}

Figure~\ref{fig:memory-scaling} measures what follows from this on one backbone. \method{} is the cheapest interface at every length we tested and handles $100$k samples in $2.2$\,GB. OpenTSLM-SP stops at $4{,}096$ samples, because its encoder uses a learned positional table of fixed length: longer inputs are not expensive, they are unrepresentable. OpenTSLM-Flamingo holds its token count nearly constant through a perceiver, but carries a fixed overhead of about $2.2$\,GB, four times ours at short lengths.

\subsection{Training on A Series}
Token count also sets the cost of a training step. Every number in Table~\ref{tab:cost} is measured at batch size $1$, the Phase-2 batch, with each interface built at the geometry of the checkpoint we report and optimising its own trainable set; a step is one forward, backward and optimiser pass, averaged over $12$ steps after warmup.

\begin{table}[t]
\centering
\setlength{\tabcolsep}{5pt}
\caption{\textbf{Phase-2 cost per example on one GPU} (Gemma3-270m, batch $1$). \emph{tok} is the context the series occupies, constant for Flamingo, whose perceiver compresses to $64$ latents; \emph{s/ex} is one training step and \emph{GB} peak training memory.}
\label{tab:cost}
\footnotesize
\begin{tabular}{lcccccc}
\toprule
& \multicolumn{3}{c}{Sleep-EDF ($1\times1500$)} & \multicolumn{3}{c}{ECG-QA ($12\times1000$)} \\
\cmidrule(lr){2-4}\cmidrule(lr){5-7}
Interface & tok & s/ex & GB & tok & s/ex & GB \\
\midrule
\method{} (ours)   & $47$   & $0.26$ & $\mathbf{0.63}$ & $384$  & $\mathbf{0.28}$ & $\mathbf{0.65}$ \\
OpenTSLM-SP        & $375$  & $\mathbf{0.19}$ & $2.44$ & $3000$ & $0.85$ & $16.88$ \\
OpenTSLM-Flamingo  & const. & $0.22$ & $9.14$ & const. & $0.23$ & $9.11$ \\
Tokenized          & $7239$ & $3.48$ & $22.78$ & $7582$ & $3.76$ & $23.82$ \\
\bottomrule
\end{tabular}
\end{table}

What Table~\ref{tab:cost} shows is a scaling property. Going from one channel of $1{,}500$ samples to twelve channels of $1{,}000$ multiplies our step time by $1.05$ and our peak memory by $1.03$. The same change costs OpenTSLM-SP $4.4\times$ in time and $6.9\times$ in memory, because a patch length of four turns twelve leads into $3{,}000$ soft tokens where we use $384$. On the twelve-lead column we therefore train $3.1\times$ faster than OpenTSLM-SP in $1/26$ of its memory, and $13.6\times$ faster than the serialized interface in $1/37$ of its memory. OpenTSLM-SP is faster on the short single-channel column ($0.19$\,s against $0.26$\,s), and that ordering reverses as soon as the input is multichannel, which is the regime the interface exists for. OpenTSLM-Flamingo is input-invariant too, since its perceiver fixes the token count, but it pays $14\times$ our training memory and two orders of magnitude more trainable parameters. The serialized baseline looks invariant only because it throws data away: its harness decimates to $1{,}200$ values across all channels, so the twelve-lead sample reaches the model with $90\%$ of its samples gone and still costs $23.8$\,GB to train on.

Generation latency separates the baselines far less, and we report it so the efficiency claim is not over-read. At a $200$-token budget every interface falls between $8.2$ and $11.1$ seconds per sample, and the spread does not change between the two input geometries: decoding is dominated by $200$ sequential steps, so a $7{,}239$-token prompt and a $47$-token prompt finish within $15\%$ of each other. The interface decides memory and reach at inference, not latency.

\section{Computational Resources}
\label{app:compute}

Table~\ref{tab:runs} reports what the Phase~1 predictive-grounding runs consumed.

\begin{table}[t]
\centering
\caption{\textbf{Phase~1 training runs}. Wall-clock time, allocation, and GPU-hours are from scheduler records. An epoch is one pass through the weighted sample drawn from the four corpora, not through every available example. Peak VRAM is the per-device maximum from a single-process V100 run at the listed per-device batch size.}
\label{tab:runs}
\small
\setlength{\tabcolsep}{4.5pt}
\begin{tabular}{@{}lcccccrrc@{}}
\toprule
Backbone & Encoder & GPUs & \multicolumn{2}{c}{Batch} & Peak & Wall clock & GPU-h & Epochs \\
\cmidrule(lr){4-5}
 & & & /device & eff. & VRAM & & & \\
\midrule
Gemma3-270M & Transformer Decoder         & $2\times4$ & $16$ & $128$ & $31.3$\,GB & $26$\,h   & $210$   & $0.7$ \\
Qwen3-0.6B  & Transformer Decoder & $2\times4$ & $8$  & $64$  & $9.1$\,GB  & $5.0$\,d  & $960$   & $2.1$ \\
Gemma3-1B   & Transformer Decoder & $4\times4$ & $8$  & $128$ & $15.2$\,GB & $10.0$\,d & $3{,}840$ & $6.2$ \\
\midrule
\multicolumn{7}{@{}l}{Total} & $5{,}010$ & \\
\bottomrule
\end{tabular}
\end{table}

\subsection{Hardware and Precision}
All runs used NVIDIA V100 accelerators in four-GPU nodes, with data parallelism only, so each device held a complete model replica. V100s have no native bfloat16 arithmetic, so we used fp16 automatic mixed precision with gradient scaling.

\subsection{Training Time} Each Phase~1 run is trained for at most ten days under the same early-stopping protocol. Validation loss is evaluated at regular intervals, and training stops when it does not improve for five consecutive validation checks. We retain the checkpoint with the lowest validation loss rather than the final checkpoint. Table~\ref{tab:runs} reports the actual wall-clock time and number of optimizer updates.

\subsection{Vocabulary-dependent Memory}
Peak VRAM is not monotonic in backbone size: Gemma3-270M uses $31.3$\,GB where Gemma3-1B uses $15.2$\,GB. The cross-entropy logit tensor explains almost all of it, since its shape is $\text{batch}\times\text{length}\times|\mathcal{V}|$. The two Gemma backbones share a $262{,}144$-token vocabulary but run at per-device batches of $16$ and $8$, so the smaller one materialises twice the logits at a given sequence length, while Qwen3-0.6B pairs batch $8$ with a smaller vocabulary of $151{,}936$. Taking Qwen3-0.6B as the reference, $\text{batch}\times|\mathcal{V}|$ predicts $1{:}1.73{:}3.45$ for Qwen3-0.6B, Gemma3-1B, and Gemma3-270M against measured ratios of $1{:}1.67{:}3.44$. Weights and activations contribute as well, but vocabulary size and per-device batch dominate. These single-process measurements omit the gradient buckets that data parallelism introduces, and the Gemma3-270M measurement enables the two-dimensional position-ID path that the original run predates, whose position tensor is negligible beside the logits.

\subsection{Configurations Across Backbones}
Gemma3-270M uses the time series encoder with LoRA adapters, training $3.17$M parameters. Gemma3-1B and Qwen3-0.6B use the transformer encoder with LoRA adapters, training $14.73$M and $12.65$M, of which $6.52$M and $5.05$M are adapter weight, and halve the per-device batch to $8$. Gemma3-1B uses twice the replicas of Qwen3-0.6B, which restores its effective batch to $128$. These totals include the next-patch decoder, which Phase~2 discards; the Params column of Table~\ref{tab:main} reports the Phase-2 trainable set and is correspondingly smaller.

\section{Prompt Templates}
\label{sec:app:prompts}

\subsection{Phase 1}
\label{sec:app:prompt_phase1}

In phase 1, we use the caption on the channel block (Figure~\ref{fig:prompt-phase1-forecast}), giving the modality, the number of channels and time steps, and the mean and standard deviation of the raw series. Training uses the next-patch regression of Eq.~\ref{eq:pred}, so no token in this prompt is a target. A forecasting batch therefore produces no cross-entropy term, which is why each Phase~1 batch is drawn from a single source.

\begin{figure}[H]
\begin{promptbox}{Forecasting (electricity demand and the synthetic corpus)}
\footnotesize\ttfamily
\ppart{pre-prompt}
\rmfamily\upshape\itshape (empty)\ttfamily\upshape\\[4pt]
\ppart{channel block}
This is a univariate time series with 1 channel and 1024 time steps. It has mean 6084.2410 and std 1174.1908. \tsblock\\[4pt]
\ppart{post-prompt}
<eos>
\end{promptbox}
\caption{Phase-1 forecasting prompt, shown for a $1{,}024$-step window of electricity demand. A multivariate item from the synthetic corpus takes the same form with \texttt{multivariate}, its channel count, and one block per channel. The answer field is the end-of-sequence token; the training signal is the next-patch regression rather than any token in this prompt.}
\label{fig:prompt-phase1-forecast}
\end{figure}

The two language corpora in Phase~1 follow the three-part schema of
\S\ref{sec:app:prompts_phase2}. Time-series QA places the item's question in the
pre-prompt and \texttt{Predict the \{task\} Answer:} in the post-prompt, with the
answer letter as the target. M4 captioning uses \texttt{You are an expert in time
series analysis.} followed by \texttt{Please generate a detailed caption for this
time-series, describing it as accurately as possible.}, with the reference
caption as the target.

\subsection{Phase 2}
\label{sec:app:prompts_phase2}

Every corpus uses the same three-part schema: a \emph{pre-prompt} stating the task and instructions, one captioned block per channel, and a \emph{post-prompt} fixing the answer format. In \method{} and the soft-prompt baselines each block holds the encoder's soft tokens, delimited by \texttt{<ts>}\,\ldots\,\texttt{</ts>}; in the serialized baseline the same position holds the decimated values as text. Each caption names the channel and gives its mean and standard deviation, so normalisation statistics reach every interface as text. Figures~\ref{fig:prompt-har}--\ref{fig:prompt-timemqa} reproduce the templates verbatim; ellipses mark truncation only.

\begin{figure}[H]
\begin{promptbox}{HAR}
\footnotesize\ttfamily
\ppart{pre-prompt}
You are given accelerometer data in all three dimensions. Your task is to classify the activity based on analysis of the data.\\[2pt]
Instructions:\\
- Begin by analyzing the time series without assuming a specific label.\\
- Think step-by-step about what the observed patterns suggest regarding movement intensity and behavior.\\
- Write your rationale as a single, natural paragraph --- do not use bullet points, numbered steps, or section headings.\\
- Do **not** mention any class label until the final sentence.\\[2pt]
Possible activity labels are:\\
biking, lying, running, sitting, standing, walking, walking\_down, walking\_up.\\[2pt]
- Make sure that your last word is the answer. You MUST end your response with "Answer: "\\[4pt]
\ppart{channel blocks (one per axis)}
The following is the accelerometer data on the x-axis, it has mean -3.2434 and std 0.0474: \tsblock\\
The following is the accelerometer data on the y-axis, it has mean 2.3132 and std 0.0550: \tsblock\\
The following is the accelerometer data on the z-axis, it has mean 9.2017 and std 0.0639: \tsblock\\[4pt]
\ppart{post-prompt}
Rationale:
\end{promptbox}
\caption{Prompt template for HAR.}
\label{fig:prompt-har}
\end{figure}

\begin{figure}[H]
\begin{promptbox}{PAMAP2}
\footnotesize\ttfamily
\ppart{pre-prompt}
You are given accelerometer data in all three dimensions. Your task is to classify the activity based on analysis of the data.\\[2pt]
Instructions:\\
- Begin by analyzing the time series without assuming a specific label.\\
- Think step-by-step about what the observed patterns suggest regarding movement intensity and behavior.\\
- Write your rationale as a single, natural paragraph --- do not use bullet points, numbered steps, or section headings.\\
- Do **not** mention any class label until the final sentence.\\[2pt]
Possible activity labels are:\\
lying, sitting, standing, walking, running, cycling, nordic walking, watching TV, computer work, car driving, ascending stairs, descending stairs, vacuum cleaning, ironing, folding laundry, house cleaning, playing soccer, rope jumping.\\[2pt]
- Make sure that your last word is the answer. You MUST end your response with "Answer: "\\[4pt]
\ppart{channel blocks (one per axis)}
The following is the accelerometer data on the x-axis, it has mean -3.2434 and std 0.0474: \tsblock\\
The following is the accelerometer data on the y-axis, it has mean 2.3132 and std 0.0550: \tsblock\\
The following is the accelerometer data on the z-axis, it has mean 9.2017 and std 0.0639: \tsblock\\[4pt]
\ppart{post-prompt}
Rationale:
\end{promptbox}
\caption{Prompt template for PAMAP2. The template matches HAR except for the eighteen-way label set.}
\label{fig:prompt-pamap2}
\end{figure}

\begin{figure}[H]
\begin{promptbox}{Sleep-EDF}
\footnotesize\ttfamily
\ppart{pre-prompt}
You are given a 30-second EEG time series segment. Your task is to classify the sleep stage based on analysis of the data.\\[2pt]
Instructions:\\
- Analyze the data objectively without presuming a particular label.\\
- Reason carefully and methodically about what the signal patterns suggest regarding sleep stage.\\
- Write your reasoning as a single, coherent paragraph. Do not use bullet points, lists, or section headers.\\
- Only reveal the correct class at the very end.\\
- Never state that you are uncertain or unable to classify the data. You must always provide a rationale and a final answer.\\[4pt]
\ppart{channel block}
This is the EEG signal, it has mean 0.0021 and std 0.9873: \tsblock\\[4pt]
\ppart{post-prompt}
Possible sleep stages are:\\
Wake, Non-REM stage 1, Non-REM stage 2, Non-REM stage 3, Non-REM stage 4, REM sleep\\[2pt]
- Please now write your rationale. Make sure that your last word is the answer. You MUST end your response with "Answer: "
\end{promptbox}
\caption{Prompt template for Sleep-EDF. Unlike the accelerometer corpora, the class list is placed in the post-prompt.}
\label{fig:prompt-sleep}
\end{figure}

\begin{figure}[H]
\begin{promptbox}{ECG-QA}
\footnotesize\ttfamily
\ppart{pre-prompt}
You are an expert cardiologist analyzing an ECG (electrocardiogram).\\
Clinical Context: 82-year-old male patient. 12-lead ECG. clinical recording. recorded with AT-6 C 5.8. Signal quality: baseline drift noted, static noise present, burst noise present, electrode artifacts present. extra beats detected during recording. pacemaker present.\\
Your task is to examine the ECG signal and answer the following medical question:\\
Question: Does this ECG show burst noise?\\[2pt]
Instructions:\\
- Begin by analyzing the time series without assuming a specific answer.\\
- Think step-by-step about what the observed patterns suggest regarding the cardiac condition.\\
- Write your rationale as a single, natural paragraph --- do not use bullet points, numbered steps, or section headings.\\
- Do **not** mention any final answer until the very end.\\
- Consider the ECG morphology, intervals, and any abnormalities that relate to the question.\\[4pt]
\ppart{channel blocks (one per lead, twelve in total)}
This is ECG Lead I, it has mean -0.0063 and std 0.1198: \tsblock\\
\ldots\ (Leads II, III, aVR, aVL, aVF, V1--V6)\\[4pt]
\ppart{post-prompt}
Based on your analysis of the ECG data, select your answer from the following options:\\
no, yes\\
- Make sure that your last word is the answer. You MUST end your response with "Answer: "
\end{promptbox}
\caption{Prompt template for ECG-QA. The clinical-context paragraph is assembled from the record's metadata; the answer set in the post-prompt is that of the question template.}
\label{fig:prompt-ecg}
\end{figure}

\begin{figure}[H]
\begin{promptbox}{TimeMQA}
\footnotesize\ttfamily
\ppart{pre-prompt}
You are given a time series from the Healthcare domain. Answer the following question.\\
Which data segment shows more volatility? A) <the provided time series> or B) [0.71, 0.72, 0.73, 0.73, 0.73, 0.73, 0.72, 0.72, 0.71, 0.7, 0.68, 0.67]\\[4pt]
\ppart{channel block}
This is the time series, mean 0.5300 std 0.1352. \tsblock\\[4pt]
\ppart{post-prompt}
Answer:
\end{promptbox}
\caption{Prompt template for TimeMQA. The corpus is open-ended, so the post-prompt is a bare \texttt{Answer:}. In comparison questions the first segment enters through the time-series interface while the second is inlined as digits, so the model compares a soft-token series against a textual one.}
\label{fig:prompt-timemqa}
\end{figure}

\clearpage

\subsection{TSRBench Evaluation}
\label{sec:app:prompts_tsrbench}
TSRBench is evaluated zero-shot, so the prompts are the benchmark's own rather than ours: no template is tuned for \method{} and every system sees the same text. All fourteen subsets are multiple choice and share one answer instruction, which fixes the output format and separates the rationale from the graded answer; only the way the signal is introduced differs between the three templates below. Series enter at the \tsblock{} placeholders exactly as in Phase~2, so the interface is unchanged from training. Parsing takes the single letter inside \texttt{<answer>}; an item with no recoverable letter is scored incorrect.

\begin{figure}[H]
\begin{promptbox}{Standard multiple choice (twelve of the fourteen subsets)}
\footnotesize\ttfamily
\ppart{question and channel blocks}
\{question\} Here are the time series '\{name$_1$\}': \tsblock. '\{name$_2$\}': \tsblock. \ldots\\[4pt]
\ppart{answer instruction (shared by all subsets)}
Select from the options below:\\
A. \{choice A\}\\
B. \{choice B\}\\
\ldots\\
Output your reasoning process in <think> tags and final answer as a single letter in <answer> tags. Format:\\
<think>Your reasoning here (less than 2048 tokens)</think>\\
<answer>A</answer>
\end{promptbox}
\caption{Standard TSRBench template. Options are labelled \texttt{A}--\texttt{G}
in the order the benchmark stores them.}
\label{fig:prompt-tsrbench-standard}
\end{figure}

\begin{figure}[H]
\begin{promptbox}{Temporal relation reasoning}
\footnotesize\ttfamily
\ppart{as above, but with the index channel removed}
\{question\} Here are the time series '\{name$_1$\}': \tsblock. \ldots\\[2pt]
\rmfamily\upshape\itshape Any channel named ``Time Stamps'' is dropped before
the placeholders are emitted, so only measured channels are encoded; the answer
instruction is unchanged.
\end{promptbox}
\caption{Temporal relation reasoning. The subset ships an explicit timestamp
channel, which would otherwise be encoded as though it were a signal.}
\label{fig:prompt-tsrbench-temporal}
\end{figure}

\begin{figure}[H]
\begin{promptbox}{Abductive reasoning}
\footnotesize\ttfamily
\ppart{pre-prompt}
You are an expert in basketball game analysis. Your task is to perform abductive reasoning.\\
Given a sequence of past events, future events, and corresponding time series data from a game, determine the most plausible event that occurred in between to link them.\\[4pt]
\ppart{context}
--- CONTEXT ---\\
Past Events (History):\\
- \{time\}: \{event\} \ldots\\[2pt]
\ldots [A CRITICAL EVENT HAPPENED HERE] \ldots\\[2pt]
Future Events:\\
- \{time\}: \{event\} \ldots\\[2pt]
Time Series Data (Win Probability):\\
Time | Team A Win Prob | Team B Win Prob\\
\ldots\\[4pt]
\ppart{task, channel blocks, then the shared answer instruction}
--- TASK ---\\
\{question\} Here are the time series 'Team A Win Probability': \tsblock. 'Team B Win Probability': \tsblock.
\end{promptbox}
\caption{Abductive reasoning. A window of ten steps either side of the critical
moment is kept. The win-probability channels are given twice, once serialised in
the table and once through the encoder, which is the benchmark's own design and
the one setting where our interface receives the signal redundantly.}
\label{fig:prompt-tsrbench-abductive}
\end{figure}

\end{document}